%% file: main.tex
\documentclass[10pt,twocolumn,letterpaper]{article}

\usepackage[pagenumbers]{cvpr} 

\usepackage{multirow}
\usepackage{arydshln}
\usepackage{graphicx}
\usepackage{booktabs}
\usepackage{colortbl}
\usepackage{amssymb} 
\usepackage{xcolor}
\usepackage{pifont} 
\usepackage{amsmath}
\usepackage{amssymb}
\usepackage{url} 
\usepackage{bm} 
\usepackage{wrapfig}
\newcommand{\cmark}{\ding{51}}
\newcommand{\xmark}{\ding{55}}

\definecolor{bestgreen}{RGB}{152,251,152}    
\definecolor{secondyellow}{RGB}{255,255,224}  
\definecolor{thirdorange}{RGB}{255,218,185}   
\definecolor{headergray}{RGB}{240,240,240}    
\definecolor{brightred}{RGB}{255, 120, 120} %
\definecolor{lightred}{RGB}{255, 178, 178}  %
\definecolor{lighterred}{RGB}{255, 222, 222}  %
\renewcommand{\arraystretch}{1.1}


\definecolor{cvprblue}{rgb}{0.21,0.49,0.74}
\usepackage[pagebackref,breaklinks,colorlinks,allcolors=cvprblue]{hyperref}

\def\paperID{} 
\def\confName{CVPR}
\def\confYear{2025}

\title{Online Language Splatting}

\author{
    Saimouli Katragadda$^{1}$,\, Cho-Ying Wu$^{2}$,\, Yuliang Guo$^{2}$\thanks{Corresponding Author.} ,\, Xinyu Huang$^{2}$,\, Guoquan Huang$^{1}$,\, Liu Ren$^{2}$
    \\
    {\small $^{1}$University of Delaware, 
    $^{2}$Bosch Research North America \& Bosch Center for Artificial Intelligence (BCAI)}
    \\
    {\small\texttt{\{saimouli,ghuang\}@udel.edu}} \\
    {\small\texttt{\{Cho-Ying.Wu,yuliang.guo2,xinyu.huang,liu.ren\}@us.bosch.edu}}\\
    {\small\url{https://saimouli.github.io/onlineLang}}
}

\begin{document}

\makeatletter
\let\@oldmaketitle\@maketitle
\renewcommand{\@maketitle}{\@oldmaketitle
\vspace{-5pt}
\centering\hspace*{2.0cm}\includegraphics[trim=0mm 0mm 0mm 0mm, clip, width=0.8\linewidth]{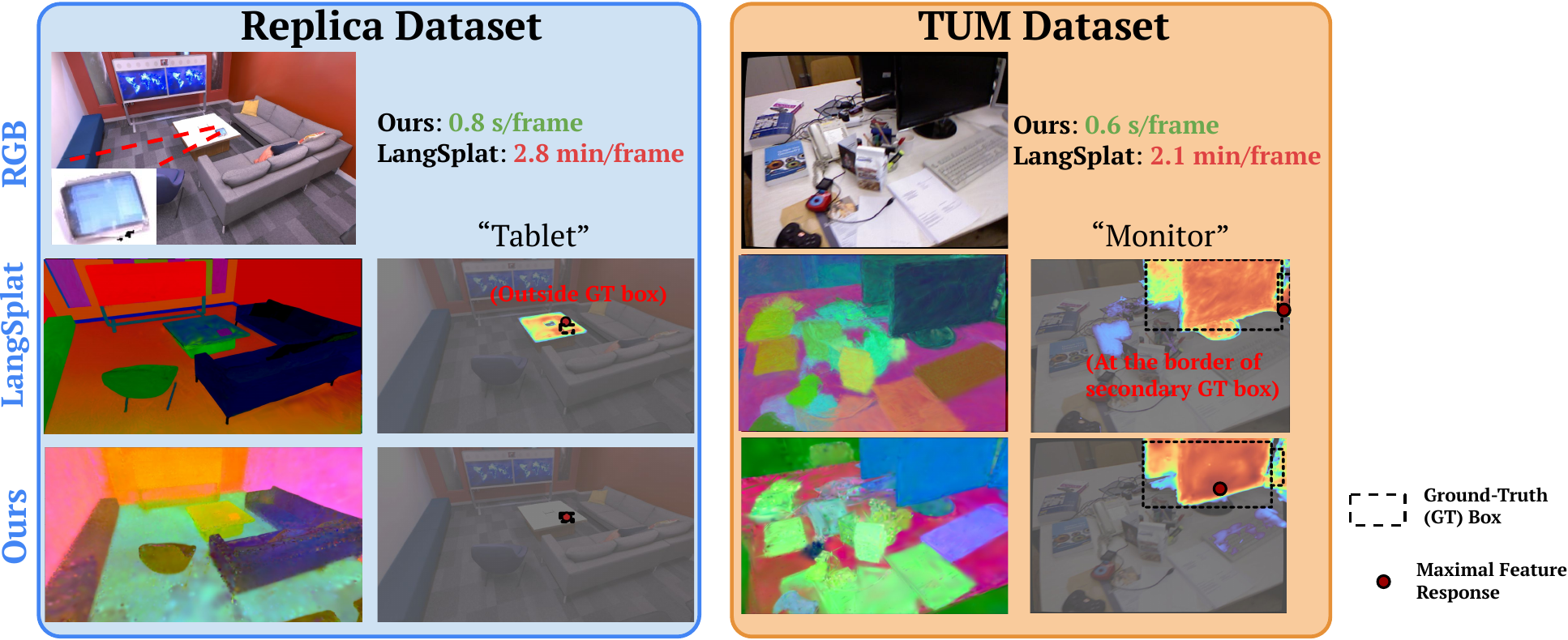}
\vspace{-5pt}
\captionof{figure}{We introduce an \textbf{online language splatting} framework that incrementally constructs a 3D Gaussian-based language feature map using 3D Gaussian Splatting within a SLAM framework. Results are compared to the offline state-of-the-art LangSplat~\cite{qin2024langsplat} across two datasets, presented in two panels. For each panel, the left column displays rendered language feature maps, and the right column shows target localization results. Our method not only outperforms in localization accuracy, but also achieve more than $40\times$ improved efficiency.
}
\vspace{8pt}
\label{teaser}}

\maketitle


\begin{abstract}
To enable AI agents to interact seamlessly with both humans and 3D environments, they must not only perceive the 3D world accurately but also align human language with 3D spatial representations. While prior work has made significant progress by integrating language features into geometrically detailed 3D scene representations using 3D Gaussian Splatting (GS), these approaches rely on computationally intensive offline preprocessing of language features for each input image, limiting adaptability to new environments.
In this work, we introduce \textbf{Online Language Splatting}, the first framework to achieve online, near real-time, open-vocabulary language mapping within a 3DGS-SLAM system without requiring pre-generated language features. 
The key challenge lies in efficiently fusing high-dimensional language features into 3D representations while balancing the computation speed, memory usage, rendering quality and open-vocabulary capability. 
To this end, we innovatively design: (1) a high-resolution CLIP embedding module capable of generating detailed language feature maps in 18ms per frame, (2) a two-stage online auto-encoder that compresses 768-dimensional CLIP features to 15 dimensions while preserving open-vocabulary capabilities, and (3) a color-language disentangled optimization approach to improve rendering quality.
Experimental results show that our online method not only surpasses the state-of-the-art offline methods in accuracy but also achieves more than $40\times$ efficiency boost, demonstrating the potential for dynamic and interactive AI applications.
\vspace{-20pt}
\end{abstract}


\section{Introduction}
\label{sec:intro}

Radiance Fields~\cite{rosinol2023nerf,journals/tog/MullerESK22:instantnerf,Kerbl:etal:SIGGRAPH2023:GaussianSplatting} have emerged as a transformative technology for 3D scene representation. Among them, 3D Gaussian Splatting (GS)~\cite{Kerbl:etal:SIGGRAPH2023:GaussianSplatting} has become particularly popular due to its high rendering quality and efficiency in differentiable rendering research. While radiance fields provide detailed geometric and textured 3D representations for photorealistic image rendering, they lack the semantic information necessary for interaction with humans.

The integration of language features into 3D scene representations has recently enabled open-vocabulary language queries, improving both interpretability and interactivity in human-computer interaction~\cite{kerr2023lerf,qin2024langsplat, shi2024language,zhou2024feature3dgs}. For example, LangSplat~\cite{qin2024langsplat} embeds CLIP-based language features~\cite{radford2021learning} into 3DGS, including both RGB and language channels per Gaussian. However, existing Lang-GS methods typically rely on computationally intensive preprocessing to generate pixel-wise language features using multimodal foundation models like SAM+CLIP, which can require minutes per frame. This substantial computational overhead limits their applicability to offline scenarios, where language features must be precomputed for each frame.

While offline language mapping is sufficient for static, predefined environments, many real-world applications demand immediate scene understanding. For instance, a service robot entering a new environment must quickly perceive the 3D surroundings to follow commands, and augmented reality (AR) systems need to deliver instant, interactive feedback as users explore new spaces. Recent advancements in combining Gaussian Splatting with online mapping~\cite{matsuki2024gaussian, keetha2024splatam, deng2024compact, yan2024gs, guo2024motiongs} have enabled detailed geometric and textured maps to be created in near real-time. However, these approaches do not incorporate language features, focusing solely on geometry and texture. Alternatively, methods that use pre-annotated ground-truth semantic maps~\cite{li2024gs3lam,ji2024neds,li2025sgs} simplify the problem but are limited to closed-vocabulary settings, lacking the flexibility required for open-vocabulary commanding.

The key challenge in online 3D language mapping lies in efficiently integrating language features into 3D representations while preserving open-vocabulary capabilities.
To address this, we introduce \textbf{Online Language Splatting}, the first framework to achieve near real-time, open-vocabulary 3D language mapping within a SLAM-GS system, eliminating the need for pre-generated language maps. Fig.~\ref{teaser} illustrates the proposed framework.
In particular, our method addresses three core sub-challenges:
(1) \textbf{Real-time High-Resolution CLIP Embedding:} Since offline, segment-centric CLIP feature preparation is a major runtime bottleneck, we replace it with a single-stage CLIP embedding and a Super-Resolution Decoder (SRD) module, enabling the generation of detailed, pixel-aligned CLIP maps in 18 ms per frame (Sec.~\ref{sec:hr_module}).
(2) \textbf{Open-Vocabulary-Preserving Feature Compression in Novel Scenes:} Unlike offline methods, which allow feature compression modules to be trained on the test scene, online methods rely on a pre-trained feature compressor to operate directly on unseen data. However, due to domain gaps, a single pre-trained autoencoder may struggle to maintain open-vocabulary capabilities when compressing CLIP features for online mapping.
To address this online-specific generalization challenge, we introduce a two-stage autoencoder, where the second stage, an Online-Learned AutoEncoder (OLAE), dynamically adapts to the dominant data variance of the current scene. This further reduces feature dimensions while preserving critical information (Sec.~\ref{sec:online_compresser}).
(3) \textbf{Color-Language Optimization:} Existing Lang-GS systems jointly optimized color and language using the same GS parameters, but these modalities inherently prefer distinct GS parameters (see Fig.~\ref{fig:RGBL_Gaussain}). Prior work~\cite{qin2024langsplat, zhou2024feature3dgs,shi2024language} jointly optimized RGB and language using shared GS parameters, which failed to achieve optimal performance for either modality.
To address this, we disentangle RGB and language backpropagation paths, designing a set of separate GS parameters to effectively render high-quality outputs for both modalities (Sec.~\ref{sec:disentangle}).

Building on these designs, our extensive experiments demonstrate that our approach not only surpasses prior state-of-the-art (SoTA) offline Lang-GS methods in text-queried 2D and 3D object localization and segmentation but also delivers a $40\times$ to $200\times$ boost in efficiency.

In summary, the main contributions of this paper include:
\begin{itemize}
\item We introduce the first near real-time, open-vocabulary, online language splatting framework, enabling flexible interaction with human language.
\item We tackle key challenges in online language splatting by proposing a real-time high-resolution CLIP embedding, an open-vocabulary-preserving feature compressor, and a color-language disentangled optimization strategy.
\item Through comprehensive evaluation, we demonstrate that our method outperforms prior state-of-the-art offline approaches across most of key metrics while achieving over 40× efficiency gains.
\end{itemize}

\vspace{-5pt}
\section{Related Work}
\label{sec:related}

\begin{figure*}[t]
    \centering
    \centering
    \includegraphics[width=0.98\linewidth]
    {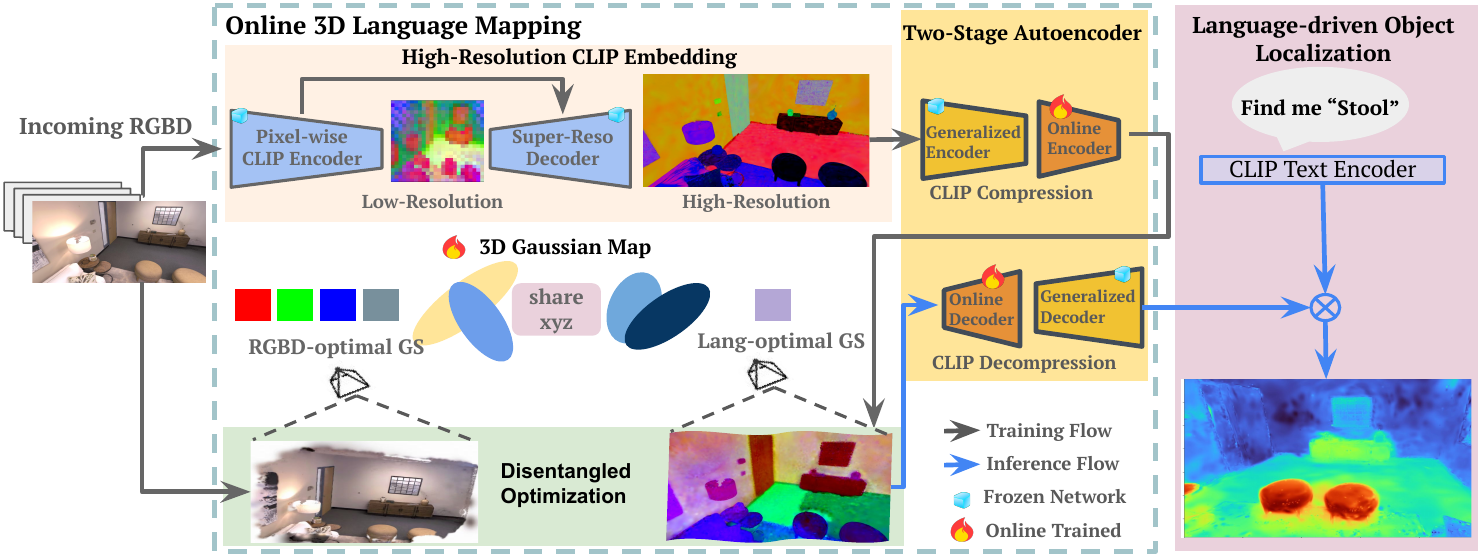} 
    \vspace{-5pt}
    \caption{
    \textbf{Online Language Splatting Pipeline.} Our pipeline integrates 3D Gaussian Splatting with SLAM, using 3D Gaussians as the sole mapping elements.
    \textbf{Left:} During training, raw images are processed through a High-Resolution (HR) CLIP embedding module, which generates HR language features in real-time. These features are compressed via a two-stage CLIP compression module into low-dimensional maps for efficient optimization while preserving open-vocabulary capabilities. RGB and language parameters are optimized separately through disentangled optimization to accommodate distinct preference in update 3D Gaussian map.
    \textbf{Right:} At inference, the rendered low-dimensional language map undergoes a two-stage decoding process to reconstruct the full-resolution CLIP feature map, enabling open-vocabulary queries to locate target objects. 
    }
    \vspace{-15pt}
    \label{fig:overview}
\end{figure*}

\vspace{-5pt}
\subsection{SLAM with Differentiable Rendering}
\vspace{-6pt}
Dense visual SLAM builds 3D maps in an online fashion, typically using classical representations such as voxel grids~\cite{dai2017bundlefusion, newcombe2011kinectfusion, whelan2015real, meilland2013unifying, mccormac2018fusion++}, Octrees~\cite{vespa2019adaptive, vespa2018efficient, xu2019mid}, or point cloud~\cite{cattaneo2022lcdnet, keller2013real, schops2019bad, campos2021orb, sandstrom2023point, whelan2015elasticfusion}. In recent years, differentiable rendering has gained popularity in SLAM, enabling the joint optimization of camera poses, maps, and implicit representations such as neural fields~\cite{sucar2021imap, zhu2022nice, rosinol2023nerf, johari2023eslam, huang2024photo, wang2023co} or explicit 3D Gaussians~\cite{matsuki2024gaussian, yan2024gs, peng2024rtg, hu2025cg, keetha2024splatam, deng2024compact, guo2024motiongs} with manageable computational overhead.
In particular, MonoGS~\cite{matsuki2024gaussian} introduces a highly efficient pipeline for online camera tracking and mapping with high-quality rendering~\cite{guo2024motiongs}, leveraging CUDA-based gradient updates to optimize camera poses. While these methods excel in rendering quality and accurate camera pose estimation, they focus solely on geometric and photometric optimization. 
We emphasize that our approach complements existing SLAM-GS methods~\cite{keetha2024splatam,peng2024rtg,matsuki2024gaussian} by introducing a novel capability: 3D open-vocabulary language mapping, extending the utility of SLAM-GS systems for more interactive applications. \textcolor{black}{For further details refer to \cite{chen2025survey3dgaussiansplatting}}.

\subsection{Language 3D Gaussian Splatting}

Lang-GS methods~\cite{qin2024langsplat,zhou2024feature3dgs,shi2024language} have recently emerged to integrate language mapping into the GS framework. Given the effectiveness of GS, embedding both language features and RGB channels into 3D Gaussians has been shown to outperform previous NeRF-based methods~\cite{kerr2023lerf} in both accuracy and rendering efficiency. However, most of these methods rely on offline-prepared language maps generated by SAM+CLIP, leading to significant processing time for open-vocabulary segmentation.
Some GS-SLAM methods~\cite{li2024gs3lam,ji2024neds,li2025sgs} support semantic map rendering but simplify the task by using dataset-provided semantic maps as ground truth. This approach limits them to closed vocabularies, contradicts the online nature of SLAM, and reduces adaptability to new scenes.
In contrast, our method is the first to achieve online open-vocabulary 3D language mapping, enabling seamless adaptation to novel environments.

\subsection{Open-Vocabulary Detection and Segmentation}
\vspace{-4pt}
Open-Vocabulary Detection (OV-D) and Segmentation (OV-S) have gained traction with the advent of CLIP~\cite{radford2021learning} and large vision-language models, enabling natural language prompts for querying. OV-D typically employs a large backbone encoder and probing heads to predict bounding boxes and classes based on cosine similarity with text embeddings~\cite{minderer2022simple, minderer2024scaling,Wang_2023_ICCV,Cheng_2024_CVPR,kuoopen, liu2023grounding}. OV-S is more challenging, requiring fine-grained masks and pixel-level semantics.
The encoders for vision-language models generally suffer from limited bottleneck feature resolution~\cite{radford2021learning,oquab2023dinov2}. Many OV-S approaches~\cite{zhang2023simple, qin2024langsplat,Liang_2023_CVPR,wang2024sam, yuan2025open, ren2024grounded,Sun_2024_CVPR} use class-agnostic mask generation via proposal networks or SAM~\cite{kirillov2023segment}, followed by vision-language model processing on proposal regions. However, these methods are computationally intensive and unsuitable for online applications. As a consequence, offline Lang-GS methods~\cite{qin2024langsplat,zhou2024feature3dgs,shi2024language} adopting SAM-based CLIP input typically takes several minutes per image on high-end GPUs to label language field ground-truth.
In contrast, we adopt a highly efficient OV-S encoding approach~\cite{xie2024sed} that directly embeds CLIP features into the network's bottleneck feature map at a low spatial resolution. To overcome the spatial resolution for precise 3D language mapping, our proposed SRD module  not only reconstructs high-resolution CLIP maps but  operates in real-time.


\section{Preliminaries}
\label{sec:prelim}
\vspace{-4pt}
\paragraph{3DGS and Rendering}
In 3D Guassian fields, each Gaussian $\mathcal{G}_i, i \in [1, \mathcal{N}]$ is represented by its 3D world-coordinate positions $\bm{\mu}_i \in \mathbb{R}^3$, covariance matrix $\bm\Sigma_i \in \mathbb{R}^{3\times3}$, colors $\bm{c}_i \in \mathbb{R}^3$, and opacity $\alpha_i \in \mathbb{R}$. 
We drop spherical harmonics the same as in the prior online 3DGS~\cite{keetha2024splatam, matsuki2024gaussian, yan2024gs}.
%
The pixel color $C$ is rendered by front-to-back composition of overlapping Gaussians sorted by depth:
\vspace{-7pt}
\begin{equation}
\small
      C = \sum_{i\in \mathcal{N}} \bm{c}_i \alpha_i      \prod_{j=1}^{i-1} (1-\alpha_j)
      =: \sum_{i\in \mathcal{N}} \bm{c}_i \alpha_i  T
\label{eq:blending}
\end{equation}
where $T$ denotes the transmittance.
Note that opacity $\alpha_i$ is after decay by the Gaussian function w.r.t. projected 2D Gaussian: $\bm{\mu}_{2D}=\mathbf{KP} \bm{\mu}, \: \bm\Sigma_{2D}=\mathbf{JR}\bm\Sigma \mathbf{R}^T \mathbf{J}^T$, 
where $\mathbf K$ is the camera intrinsics, $\mathbf P$ is the world-to-camera projection matrix, $\mathbf R$ is the rotation, and $\mathbf J$ is the Jacobian of the affine approximation of the projective transformation. 

\paragraph{SLAM Tracking and Keyframing}
We adopt tracking and keyframing mechanism in MonoGS~\cite{matsuki2024gaussian}. For each tracked frame, current camera pose is optimized with appearance and geometry loss.
\begin{equation}
\small
      \mathcal{L} = \lambda |C^r-C^{gt}| + (1-\lambda)|D^r-D^{gt}|,
\label{eq:main_loss}
\end{equation}
where $C^r$ and $D^r$ are rendered image and depth by alpha composition in Eq.~\eqref{eq:blending}. (For depth rendering, the color term is replaced by z-direction distance at the center of a Gaussian.)
Tracked frames are selected as keyframes within a local window after a co-visibility check, ensuring sufficient novel regions are visible in each keyframe and avoiding redundant optimization. For each keyframe, new Gaussians are created in the 3D maps with $\bm{\mu}$ initialized by back-projecting depth to cover the areas. The 3DGS parameters are optimized by the maintained keyframe window with appearance and geometry loss Eq.~\eqref{eq:main_loss} plus scale-isotropic regularization to prevent serious needle-like artifacts.

\paragraph{Multi-Channel Optimization}
Prior work (e.g., \cite{qin2024langsplat}) that embeds semantics or language features into 3DGS caches additional channels per Gaussian. The language map rendering follows alpha-blending rules:
\begin{equation}
\small
      F = \sum_{i\in \mathcal{N}} \bm{f}_i \alpha_i \prod_{j=1}^{i-1} (1-\alpha_j),
\label{eq:feature_rendering}
\end{equation}
where $\bm{f}_i \in \mathbb{R}^3$ is language feature in each Gaussian. During the backward pass, language gradients are entangled with color and depth gradients:
\begin{equation}
\small
      \frac{\partial \mathcal{L}}{\partial \alpha_i} = \frac{\partial \mathcal{L}}{\partial C} \frac{\partial C}{\partial \alpha_i} + \frac{\partial \mathcal{L}}{\partial D} \frac{\partial D}{\partial \alpha_i} + \frac{\partial \mathcal{L}}{\partial F} \frac{\partial F}{\partial \alpha_i},
\label{eq:naive_backprop}
\end{equation}
where the loss $\mathcal{L}$ is based on Eq.\eqref{eq:main_loss} with an additional L1 loss using groundtruth language maps via SAM+CLIP. 
Some works ~\cite{ji2024neds, li2025sgs, li2024gs3lam} use gradients in Eq.\eqref{eq:naive_backprop} to train with multi-modality online. 
To not let language features interfere with color Gaussians, LangSplat's offline optimization first trains color Gaussians on the whole sequence without the last gradient term, and in the second stage they use the last term to pass gradient to only train language features.


\label{sec:method}
\section{Online Language Splatting}

\begin{figure}[bt!]
    \centering
    \includegraphics[width=0.98\linewidth]{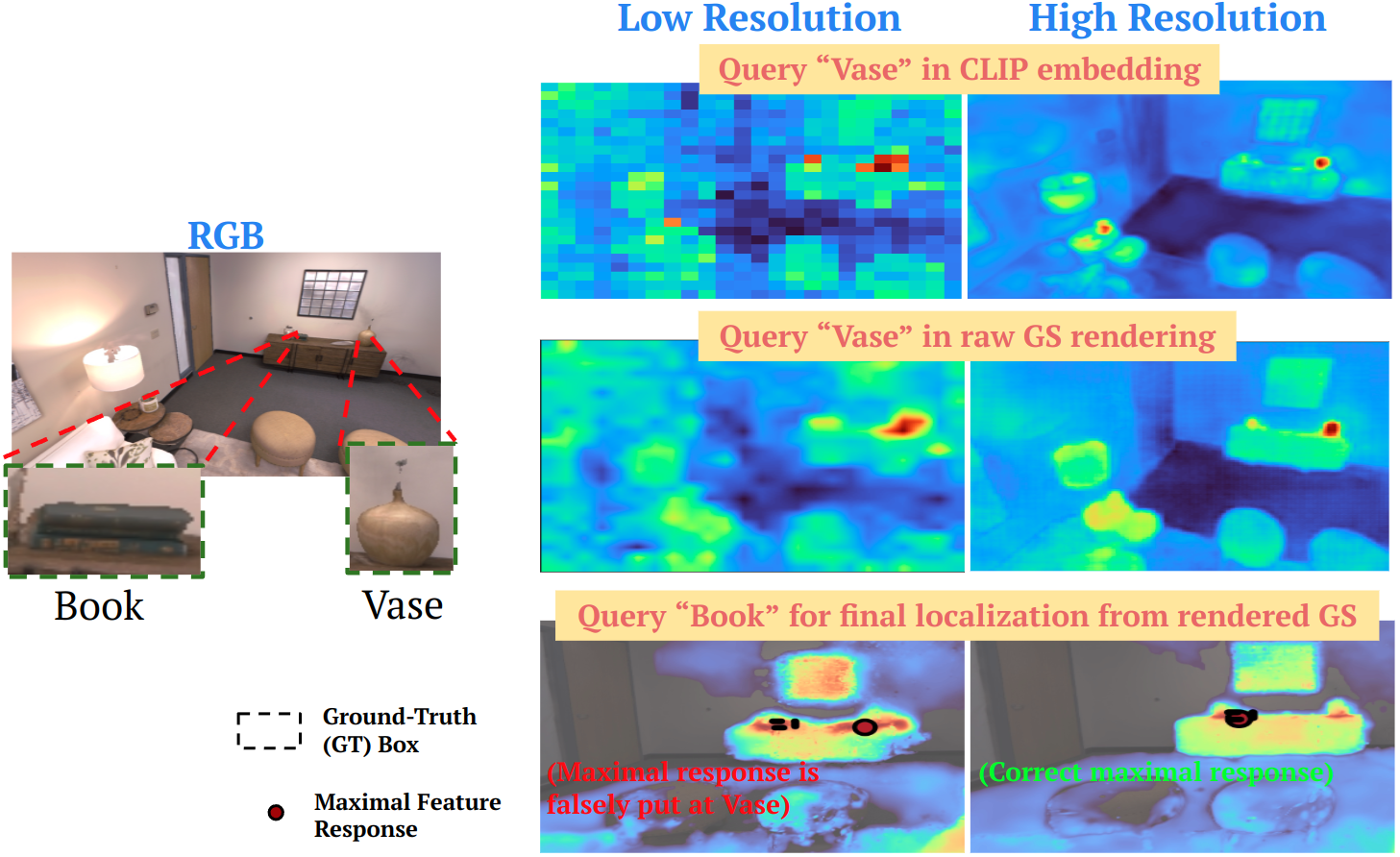}
    \caption{\textbf{Importance of Feature Map Resolution.} \textbf{Top:} Compared to the Low Resolution (LR) query heatmap from the pixel-wise encoder output (left), the High Resolution (HR) heatmap from SRD output (right) improves localization and differentiation. \textbf{Middle \& Bottom:} Query heatmaps from rendered maps after GS mapping. GS mapping from LR exhibits feature bleeding, while mapping from HR preserves structural details, better localization. }
    \label{fig:hr_result}
\end{figure}

Our approach enables near real-time, high-resolution, Open-Vocabulary (OV) language mapping within a 3D Gaussian Splatting (3DGS) framework, facilitating language-driven spatial understanding for robotics and AR applications.
As shown in Fig.~\ref{fig:overview}, our pipeline consists of three main components during the training and optimization phase, addressing the key challenges outlined in Sec.~\ref{sec:intro}. 

The pipeline begins with standard RGB-D SLAM input streams. Color images are processed through a pixel-wise CLIP encoder to generate low-resolution language features. These features, combined with hierarchical encoder outputs, are then refined by a Super-Resolution Decoder (SRD) to produce pixel-aligned, high-resolution language maps.
Next, the CLIP Compression module, implemented as a Two-Stage Autoencoder, significantly reduces the dimensionality of CLIP features for efficient online mapping while preserving essential information for OV queries. The second stage, an Online-Learned Autoencoder (OLAE), further enhances generalization to novel scenes.
Finally, Disentangled Optimization separates gradient flows for color and language, enabling independent optimization of Gaussian parameters. This improves rendering quality across both modalities.
During inference, the rendered low-dimensional language map can be passed through the Two-Stage Autoencoder to reconstruct full CLIP features, allowing OV queries for locating target objects.

%

\subsection{High-Resolution CLIP Embedding}
\label{sec:hr_module}

Unlike offline methods that require multiple passes and complex mask generation, our approach leverages a ConvNeXt-based pixel-wise CLIP Encoder~\cite{xie2024sed} to generate a coarse CLIP embedding map, which is then refined by a lightweight \textbf{Super-Resolution Decoder (SRD)} to produce dense, high-quality language maps. This design preserves conceptual integrity while enabling real-time operation. The SRD takes a coarse CLIP map along with the intermediate outputs from layers 1 and 2 of the pixel-wise encoder as inputs, progressively enhancing the CLIP feature map resolution through two convolutional upsampling blocks that align with hierarchical encoder features. The detailed architecture is illustrated in Supplemental Fig.~\ref{fig:srd_network}.


The Super-Resolution Decoder (SRD) is trained with supervision from offline high-resolution CLIP feature maps. Following a procedure similar to~\cite{qin2024langsplat}, we generate high-resolution language features to serve as training labels for our lightweight SRD.
Our training is not restricted to a specific dataset, as we only require diverse images to cover a broad range of concepts without relying on their original annotations. 
When training images encompass diverse concepts (e.g., COCO~\cite{lin2014microsoft}), and our training focuses on the simplified task of upsampling the feature map, the OV capability of the CLIP features is expected to be preserved.


The resulting high-resolution CLIP embedding module (pixel-wise CLIP Encoder + SRD) operates highly efficiently, achieving a runtime of 18 ms on an RTX-3090 GPU while using only 1.6 GB of GPU memory. The SRD sub-module contributes only 2 ms to this runtime, significantly improving feature quality with minimal overhead. These enhancements in turn result in improved accuracy and IoU (see Table~\ref{tab:replica_results:summary}, Fig.~\ref{fig:qualitative_res1}). The benefits of high-resolution CLIP maps are further illustrated at the feature level in Fig.~\ref{fig:hr_result}.

Note that our SRD design shares certain similarities with FeatUp~\cite{fu2024featup}, as both approaches focus on upsampling feature maps. However, unlike their unsupervised method, we employ a simpler and more efficient strategy by using hierarchical feature supervision to guide upsampling in a supervised manner. Our approach not only enhances accuracy for high-resolution images but also improves computational efficiency (see supplemental Fig.~\ref{fig:hr_compare} for comparison).

\subsection{Two-Stage Online CLIP Compression}
\label{sec:online_compresser}

Since CLIP features are high-dimensional (768) vectors, a key challenge is how to effectively compress them to enable real-time integration while preserving OV capabilities.

To address this, we first develop a generalized language compressor that exploits the inherent redundancy in language feature embeddings. Using diverse images from a large dataset (e.g., COCO), we train a simple autoencoder baseline with a multi-layer MLP to compress the dimensionality from 768 to a 32-dimensional code. This code size is carefully chosen to balance semantic preservation and data compression.
Due to the domain gap between the pretraining dataset and the test scenes, the output dimension cannot be too low, as excessive compression may compromise OV capabilities when applied to new domains. Supplemental Table~\ref{tab:coco_ae} provides a detailed analysis of code size selection and its impact on performance.

While the generalized language compressor effectively reduces dimensionality, the resulting code size remains too large for efficient integration into an online Lang-GS framework. To further compress the CLIP feature while preserving their OV capability, we introduce an \textbf{Online-Learned AutoEncoder (OLAE)} as a second-stage compressor, which adapts dynamically to testing scenes by compressing features into a smaller 15-dimensional code. This adaptation is based on the observation that data variance within a specific scene can often be captured by fewer dimensions, allowing less relevant directions from the generalized model to be disregarded.
The OLAE starts with an initial training phase of 200 iterations (6 ms/iter) and incrementally updates using selected keyframes. For each iteration, two additional random keyframes are incorporated, ensuring retention of previously learned features and preventing catastrophic forgetting.
By combining a generalized compressor (for broad vocabulary preservation) and an online-learned compressor (for scene adaptability), our approach maintains OV capabilities while significantly reducing memory cost, making real-time applications feasible.

\begin{figure}[bt!]
    \centering
    \includegraphics[width=0.9\linewidth]{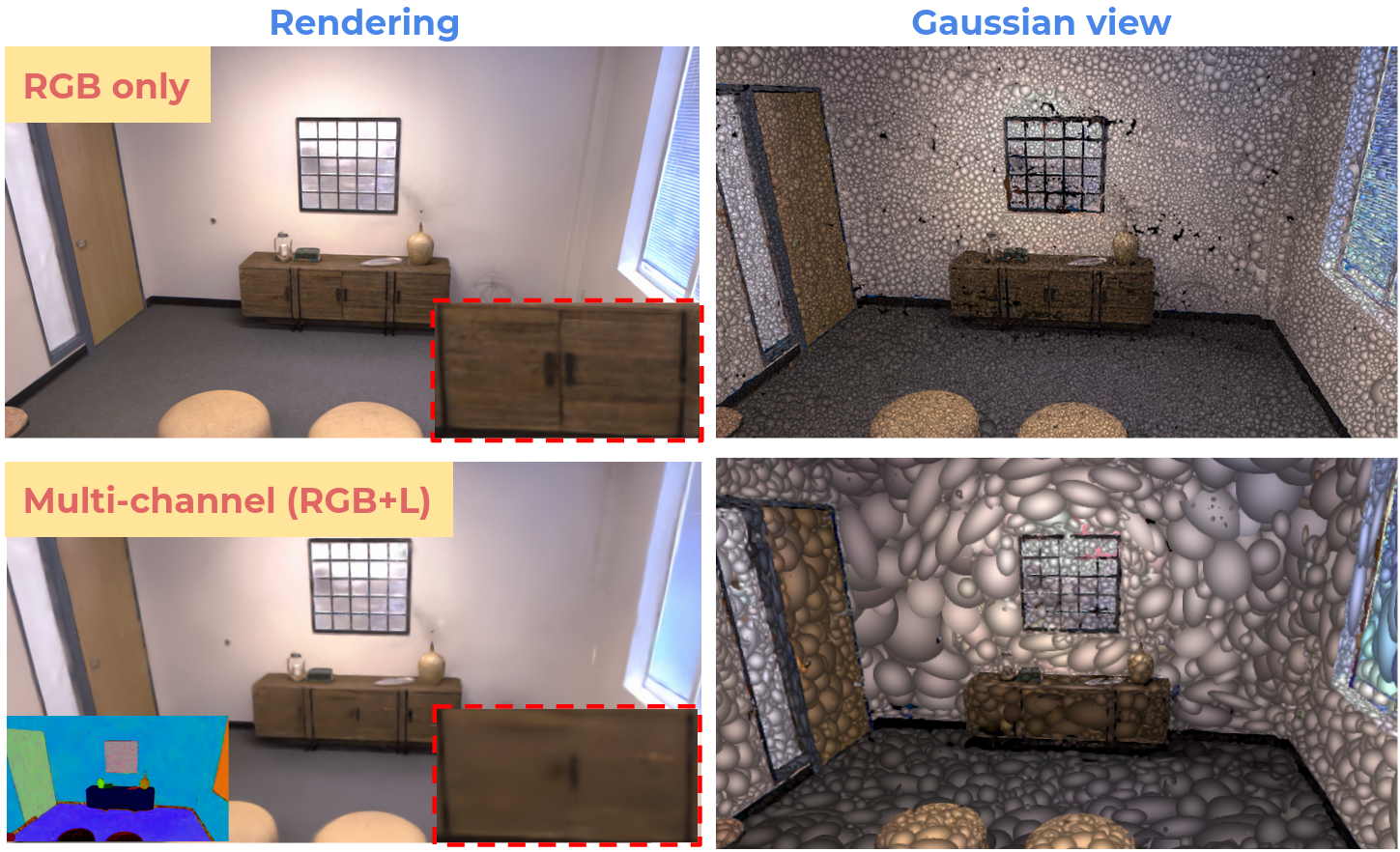}
    \vspace{-7pt}
    \caption{\textbf{Color optimization alone vs. color-language joint optimization.} Adding language channels leads to sub-optimal rendering quality with much different Gaussian maps.}
    \vspace{-16pt}
    \label{fig:RGBL_Gaussain}
\end{figure}

\subsection{Color-Language-Disentangled Optimization}
\label{sec:disentangle}

One of the key challenges in online 3DGS is how to  optimizing color and language modalities in the meantime.
Prior methods like LangSplat use additional channels in each Gaussian to represent language features, as described in Sec.~\ref{sec:prelim}. Colors and language modality share the common 3DGS parameters, including $\alpha$, $\bm{\mu}$, and $\bm\Sigma$ (or the same scale $\bm S$ and rotation $\bm R$). However, we find that sharing the common parameters will lead to suboptimal performance for both colors and language. To verify, we visualize the Gaussian fields for training colors alone and jointly optimizing RGB-L channels in Fig.~\ref{fig:RGBL_Gaussain}. The color rendering deteriorates when jointly optimizing RGB-L. Observing Fig.~\ref{fig:RGBL_Gaussain}, we find that this is because language features tend to stretch Gaussian scales and apply different rotations. Unlike color appearance with textural details, language features are homogeneous, such as wall areas are associated with the same language codes. Therefore, language Gaussians prefer much different GS parameters from RGB. 

Losses of color and language are chained to the same GS parameters by Eq.~\eqref{eq:naive_backprop}, where $\partial\mathcal{L}/\partial\alpha_i$ further back-propagates to $\bm{\mu}, \bm R, \bm S$. However, if using different sets of whole GS parameters, the number of Gaussians can be high and takes up the memory, but the key rendering contributors could still be sparse. To efficiently represent colors and language, we adopt multi-mode $\bm R$, $\bm S$, and $\alpha$ for colors and languages, and they still share the same $\bm{\mu}$ in each Gaussian to prevent Gaussian duplication. The color and language rendering become
\begin{equation}
\small
      C = \sum_{i\in \mathcal{N}} \bm{c}_i \alpha^c_i \prod_{j=1}^{i-1} (1-\alpha^c_j), \quad F = \sum_{i\in \mathcal{N}} \bm{f}_i \alpha^f_i \prod_{j=1}^{i-1} (1-\alpha^f_j), 
\label{eq:revised_rendering}
\end{equation}
and for depth rendering we use $\alpha^c$. The back-propagation becomes 
\begin{equation}
\small
      \frac{\partial \mathcal{L}}{\partial \alpha^c_i} = \frac{\partial \mathcal{L}}{\partial C} \frac{\partial C} {\partial \alpha^c_i} + \frac{\partial \mathcal{L}}{\partial D} \frac{\partial D}{\partial \alpha^c_i}, \,\,      \frac{\partial \mathcal{L}}{\partial \alpha^f_i} = \frac{\partial \mathcal{L}}{\partial F} \frac{\partial F}{\partial \alpha_i^f},
\label{eq:backprop}
\end{equation}
\begin{equation}
\small
\begin{aligned}
      \frac{\partial \mathcal{L}}{\partial \bm R_i^{c/f}} = \frac{\partial \mathcal{L}}{\partial \alpha_i^{c/f}} \frac{\partial \alpha^{c/f}_i} {\partial \bm\Sigma_i^{c/f}} \frac{\partial \bm\Sigma_i^{c/f}}{\partial \bm R_i^{c/f}},  \,\, \frac{\partial \mathcal{L}}{\partial \bm S_i^{c/f}} = \frac{\partial \mathcal{L}}{\partial \alpha_i^{c/f}} \frac{\partial \alpha^{c/f}_i} {\partial \bm \Sigma_i^{c/f}} \frac{\partial \bm\Sigma_i^{c/f}}{\partial \bm S_i^{c/f}},
\end{aligned}
\label{eq:backprop}
\end{equation}
where script $c/f$ denotes different modes. We empirically simplify the back-propagation path to $\partial \mathcal{L}/\partial \bm{\mu}_i^{2D}$ by only computing $\partial \mathcal{L}/\partial \alpha_i^c \times \partial \alpha_i^c/\partial \bm{\mu}_i^{2D}$ without language. MonoGS~\cite{matsuki2024gaussian} further computes gradients to camera poses ($\bm P$) by $\partial \bm{\mu}_i^{2D}/\partial \bm P$ and $\partial \bm\Sigma_i^{2D}/\partial \bm P$, and we also drop the language part and only use color mode's mean and covariance for camera poses. 
For co-visibility check in keyframing, we require both colors and language are with sufficient novel areas to join as keyframes. To prune a Gaussian, we also require both $\alpha_i^c$ and $\alpha_i^f$ are below a threshold. Last, 
to not let language mode learn skewed scales, we add a loss $|\bm S_i^f - \bm S^c_{i\bot}|$, where $_{\bot}$ denotes stop gradient.




\section{ Experiments}
\label{sec:experiment}
\vspace{-3pt}
\noindent\textbf{Baselines.} Since we introduce the first online Language Gaussian Splatting (Lang-GS) method, we primarily compare our approach to state-of-the-art (SoTA) offline Lang-GS methods, including LangSplat~\cite{qin2024langsplat}, Feature3DGS~\cite{zhou2024feature3dgs}, and LEGaussian~\cite{shi2024language}\footnote{LEGaussian's reported Replica results are based on re-annotated and simplified groundtruth. We re-evaluate it for fair comparison.} in text query-based object localization. Additionally, to demonstrate that integrating language mapping does not degrade SLAM system performance, we compare our method against SoTA SLAM-GS approaches in image rendering and camera pose tracking, including MonoGS~\cite{matsuki2024gaussian}, SplaTAM~\cite{keetha2024splatam}, and RTG-SLAM~\cite{peng2024rtg}.

\noindent\textbf{Datasets.} 
We conduct evaluations on two widely used datasets for the SLAM setup: the synthetic Replica~\cite{straub2019replica} and the challenging real-world TUM RGB-D~\cite{sturm12iros}, both qualitatively and quantitatively. 
In Replica, we evaluate the top 10 most frequent classes, sampling 21 frames randomly from each sequence as test frames. 
In TUM RGB-D, we manually annotate test frames to create ground-truth masks for language queries, serving as evaluation targets. In training, we utilize COCO~\cite{lin2014microsoft} and Omnidata~\cite{eftekhar2021omnidata} datasets due to their diverse range of concepts, ensuring broad generalization across various scenes and objects.


\noindent\textbf{Evaluation Metrics.} To assess object localization via text queries, we follow LangSplat to use mIoU and localization accuracy (Loc) on rendered language maps. Localization is considered successful if the highest-relevancy pixel falls within the ground-truth bounding box. To evaluate image rendering quality, we use PSNR, SSIM, and LPIPS, while camera pose tracking is assessed via ATE RMSE, following the evaluation protocol in~\cite{matsuki2024gaussian}. We exclude the post-stage global refinement for strict online requirements. Each method’s runtime is measured by per-frame processing time. Besides these known metrics, we also evaluate text-query 3D localization via Chamfer Distance (CD) and Earth Mover’s Distance (EMD) between the ground-truth and localized point sets. 
%

%

\begin{table}[t]
\centering
\footnotesize
\caption{\textbf{Comparison to Lang-GS SoTA on Replica}. Our method is compared to the SoTA Lang-GS methods on the Replica dataset in terms of image-based localization accuracy and per-frame running time. We also analyze the impact of key introduced modules, including Super-Resolution Decoder (SRD) in CLIP Embedding and Online Learning of AutoEncoder (OLAE) in feature compression. The variants without OLAE train a single AE from other scenes in Replica Dataset.}

\vspace{-5pt}
\label{tab:replica_results:summary}
\setlength{\tabcolsep}{4pt}
\begin{tabular}{l|cc|cc| c}
\hline
\rowcolor{gray!10}
{\textbf{Method}} & \multicolumn{2}{c|}{\textbf{Modules}} & \multicolumn{2}{c|}{\textbf{Query Loc.}} & {\textbf{Time}} \\
\cline{2-5}
\rowcolor{gray!10}
& SRD & OLAE & mIOU & Loc & \\
\hline
LangSplat~\cite{qin2024langsplat} & $-$ & $-$ & 0.417 & 0.720 & 2.8 min/fr \\
Feature3DGS~\cite{zhou2024feature3dgs} & $-$ & $-$ & 0.359 & 0.755 & 2.3 min/fr\\
LEGaussian~\cite{shi2024language} & $-$ & $-$ & 0.245 & 0.682 & 32 s/fr \\
\midrule
\multirow{5}{*}{Ours} 
& \xmark & \xmark & 0.400 & 0.754 & \multirow{5}{*}{\textbf{0.8 s/fr}} \\
& COCO & \xmark & 0.475 & 0.782 &  \\
& \cellcolor{thirdorange}COCO & \cellcolor{thirdorange}\cmark & \cellcolor{thirdorange}0.479 & \cellcolor{thirdorange}0.759 &  \\
& Omni & \xmark & 0.485 & 0.802 & \\
& \cellcolor{thirdorange}Omni & \cellcolor{thirdorange}\cmark & \cellcolor{thirdorange}\textbf{0.487} & \cellcolor{thirdorange}\textbf{0.826} & \\
\hline
\end{tabular}
\vspace{-2pt}
\end{table}

\input{sec/4_2_tumrgb_table}

\noindent\textbf{Implementation Details.} We utilize a pre-trained CLIP ViT-L model~\cite{radford2021learning} alongside a ConvNeXt-L-based hierarchical encoder from~\cite{xie2024sed} to extract 768-dimensional feature representations from input images~\cite{liu2022convnet}. The input to this module is an RGBD image with dimensions 640×640×3. The module processes the input and produces a feature map of size 24×24×768. Subsequently, the SRD enhances this feature map to an output resolution of 
192×192×768, maintaining the semantic context of the input data.
We train two separate SRD models: one on 7\% of the COCO dataset~\cite{lin2014microsoft} and another on 30\% of the Omnidata-Tiny dataset~\cite{eftekhar2021omnidata}. Both models are trained on four A5000 GPUs with a batch size of 12 images per GPU.
We utilize the AdamW optimizer with an initial learning rate of $2 \times 10^{-4}$ and a weight decay of $1 \times 10^{-4}$ for a total of 180 training epochs.
For the generalized auto-encoder, a 8-layer MLP architecture was used to compress language features to a code size of 32. 
For the online compressor, a 2 layer MLP with a encoder code of size 15 is trained online using a scheduler with a reduction on plateau and a threshold of $1 \times 10^{-4}$, and optimized with Adam using a learning rate of $1 \times 10^{-3}$. The online training takes 10 key frames with 200 iterations in initialization, and update 1 iteration in the consecutive frames.
To ensure a fair comparison, we upgrade LangSplat's OpenCLIP~\cite{ilharco_gabriel_2021_5143773} model to match our feature dimensions (768, up from 512), use a code size of 15 (from 3), and train the pipeline offline on the entire Replica and TUM RGBD dataset, sampling every 10th image from each sequence. 

\begin{figure}[ht]
    \centering
    \includegraphics[trim=10mm 0mm 0mm 0mm,clip, width=0.99\linewidth]{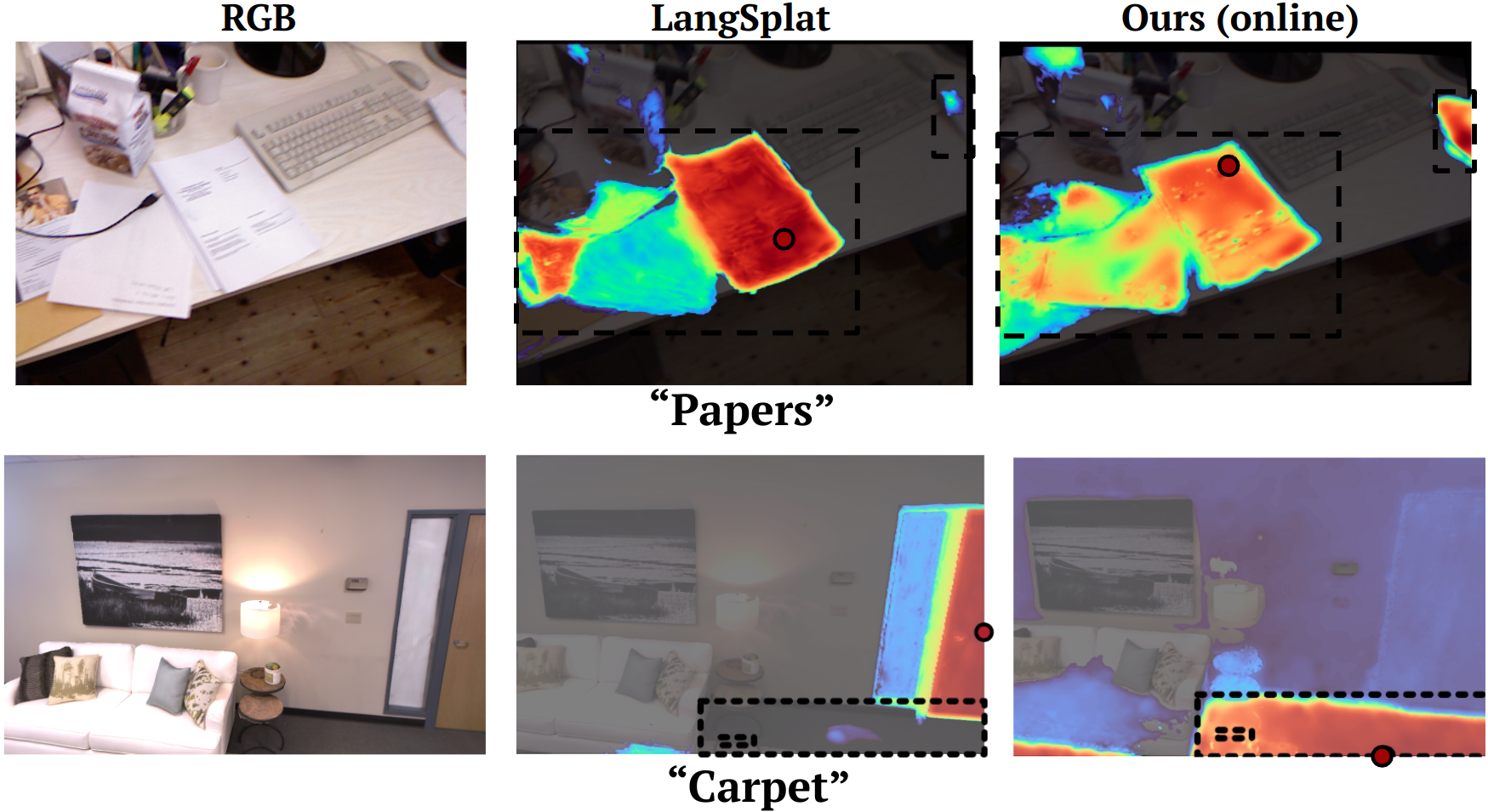}
    \vspace{-5pt}
    \caption{
    \textbf{Qualitative comparison with offline SoTA:} \textbf{Top}: On the TUM RGB-D dataset, our method successfully segments the paper in the top-right corner, which LangSplat fails to detect. \textbf{Bottom}: On the Replica dataset, we accurately localize the carpet, whereas LangSplat misidentifies a different object. Black box: ground-truth box; red dot: maximal feature response as the predicted localization.
    }
    \vspace{-10pt}
    \label{fig:qualitative_vs_sota}
\end{figure}

\subsection{Comparison with the State of the Art}

\noindent{\textbf{Comparison to Lang-GS SoTA Methods.}}
The comparison between our method and previous SoTA offline Lang-GS methods is presented in Table~\ref{tab:replica_results:summary} and Table~\ref{tab:tum_rgbd_comparison}. As observed, our method establishes a new SoTA performance on the Replica dataset, significantly surpassing offline methods, regardless of whether SRD is trained on COCO or Omnidata datasets. 
It also leads to improved localization accuracy and competitive mIoU scores on the TUM-RGBD dataset upon LangSplat. As an online method, our approach is $40\times$ to $200\times$ more efficient than SoTA offline methods. A detailed per-scene evaluation on the Replica dataset is provided in supplemental Table~\ref{tab:replica_results}.
Qualitatively, Fig.~\ref{fig:qualitative_vs_sota} shows that our method correctly identifies objects that LangSplat either misses or misidentifies. 
The ablation study of key modules are further discussed in Sec~\ref{sec:ablation}. 

On the other hand, our performance advantage on the TUM-RGBD dataset is less pronounced. This is primarily due to challenges such as motion blur and lower image quality, which complicate online camera tracking. These conditions favor offline approaches that rely on extensive global optimization (e.g., 30k iterations) of both 3D Gaussian parameters and camera poses.

\begin{table}[t]
\centering
\begin{footnotesize}
\caption{\textbf{SLAM-GS Evaluation on Replica}. Our method is evaluated against other SLAM-GS approaches based on novel view rendering quality and camera localization error (ATE in cm).}
\vspace{-7pt}
\label{tab:SOTA_comparison_slam_gs_summary}
\setlength{\tabcolsep}{3pt}  

\begin{tabular}{c|c|cccc}
\hline
\rowcolor{gray!10}
\textbf{Method}   & \textbf{Lang.} & \textbf{PSNR} $\uparrow$ & \textbf{SSIM} $\uparrow$ & \textbf{LPIPS} $\downarrow$ & \textbf{ATE (cm)} $\downarrow$ \\
\hline
SplaTAM~\cite{keetha2024splatam}  & \xmark & 33.39 & {0.968} & 0.101 & 0.392 \\
RTG-SLAM~\cite{peng2024rtg} & \xmark & 35.77 & \textbf{0.982} & 0.106 & \textbf{0.182} \\
MonoGS~\cite{matsuki2024gaussian}   & \xmark & {35.72} & 0.950 & {0.075} & 0.420 \\
Ours     & \cmark & \textbf{35.81} & 0.950 & \textbf{0.072} & 0.397 \\
\hline
\end{tabular}
\end{footnotesize}
\vspace{-5pt}
\end{table}

\noindent{\textbf{Comparison to SLAM-GS SoTA Methods.}}
We compare our method to recent SoTA SLAM-GS approaches in Table~\ref{tab:SOTA_comparison_slam_gs_summary}. As observed, despite integrating language feature mapping into a SLAM-GS system, our framework preserves novel view rendering performance and \textcolor{black}{localization is on par} to MonoGS~\cite{matsuki2024gaussian}, the SLAM framework on which we build.
Compared to other SoTA SLAM-GS methods, our approach achieves the best overall performance in novel view rendering while incorporating language mapping. RTG-SLAM achieves superior ATE due to the inclusion of additional classical SLAM modules, which increases system complexity. Per-scene results are provided in supplemental Table~\ref{tab:SOTA_comparison_slam_gs}.

\noindent{\textbf{3D Localization Evaluation}.}
For 3D localization, we first fuse multi-view language renderings into voxels by truncated signed distance field (TSDF). Each voxel size is 20cm. Then, we extract mesh by marching cubes. Each vertex is described by language features. We pass the batched point features into the CLIP decompressor, reconstruct the 768 dimensions, and compare with the mesh build by TSDF with semantic mask groundtruth.
We compare with LangSplat on the devised 3D localization protocol. Table~\ref{tab:langsplat_comp} demonstrates that our framework can run in an online setting with better 3D language mapping. Fig.~\ref{fig:3d_localization} provides visualization of the 3D localization. See the supplementary Sec~\ref{supp:3d:loc} for more results.

\vspace{-2pt}
\input{sec/4_langsplatcompare_table}
\vspace{-6pt}
\begin{figure}[h]
    \centering
    \includegraphics[width=0.98\linewidth]{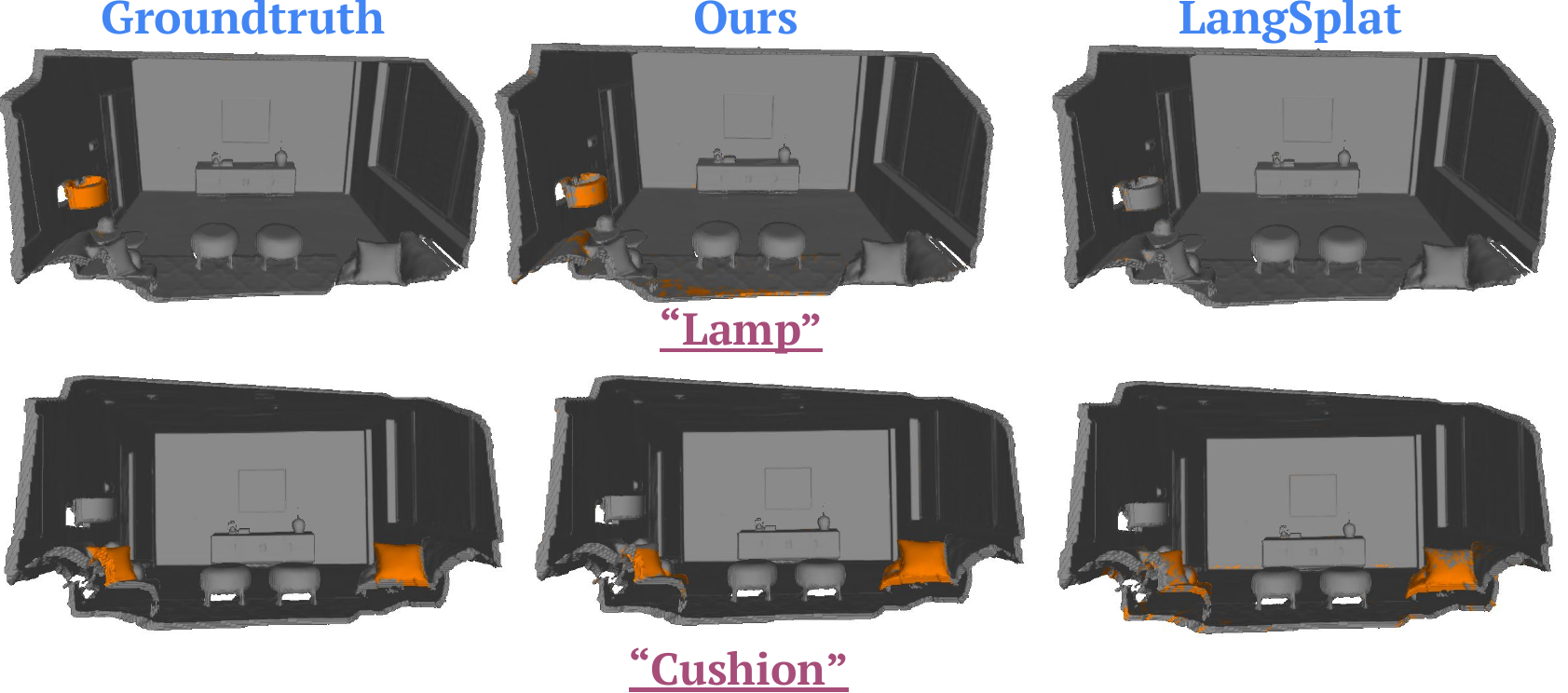}
    \vspace{-5pt}
    \caption{\textbf{Visualization of 3D localization results.}}
    \label{fig:3d_localization}
    \vspace{-14pt}
\end{figure}

\noindent{\textbf{Runtime Analysis.}} Our entire network module runs at 21ms per frame on an RTX-3090 GPU, including 15ms for CLIP encoding, 2ms for super-resolution decoding, and 6ms for online compression with online training. While the overall pipeline speed is currently bottlenecked by the MonoGS baseline—resulting in a runtime of 0.6–0.8 seconds per frame—significantly higher speeds are achievable with advancements in the SLAM-GS system. For instance, by integrating our method into Hi-SLAM \cite{zhang2024hislam2}, we achieve 7.05 FPS.
In contrast, the offline method LangSplat requires approximately 168s per frame (2.8 minutes), including 35s for SAM, 10s for post-processing, and an additional 123s per frame (amortized) for training the dense CLIP autoencoder on the testing scene. This total runtime underscores the significant computational cost of an offline approach.

\subsection{Ablation Study}
\label{sec:ablation}

\begin{figure}[ht]
    \centering
    \centering
    \includegraphics[width=0.9\linewidth]{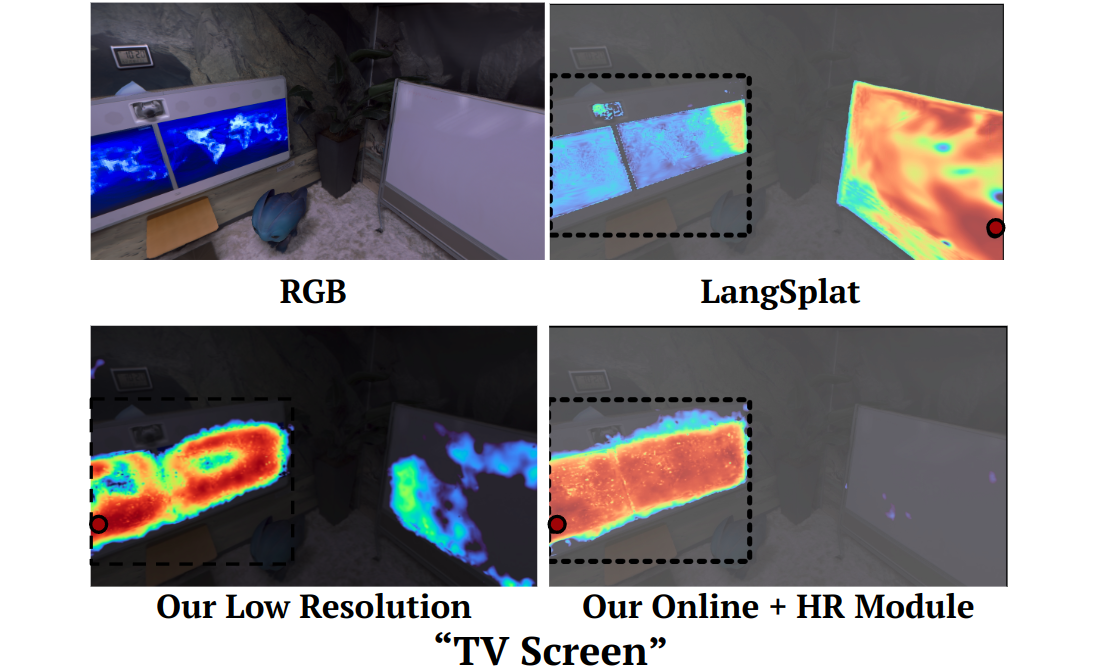}
    \vspace{-8pt}
    \caption{
    \textbf{Visual comparison in CLIP embedding:} Our High-resolution embedding allows for more complete object capture, reducing holes and resolving semantic ambiguities. The dotted black box represents the ground-truth, while the red dot indicates the maximal feature response as the predicted localization.
    }
    \label{fig:qualitative_res1}
\end{figure}

\noindent{\textbf{Super-Reso Decoder (SRD) in CLIP Embedding}} 
We analyze the impact of SRD on the Replica dataset, with results summarized in Table~\ref{tab:replica_results:summary} and individual scene results provided in supplemental Table~\ref{tab:replica_results}.
We observe that SRD significantly improves both mIoU and Loc metrics from our basic online baseline. The underlying reasons for these improvements are evident through visual comparisons in Fig.~\ref{fig:hr_result} and Fig.~\ref{fig:qualitative_res1}. From Fig.~\ref{fig:hr_result}, we can see that high-resolution language maps greatly enhance localization of small or distant objects. Fig.~\ref{fig:qualitative_res1} demonstrates improvements in semantic boundaries and even helps resolve ambiguities between visually similar but different classes. \textbf{Cross-view consistency:} In a new ablation study, we found that the generalized encoder improves cross-view consistency—without it, the average cosine similarity drops to 0.5. While frame-wise encoding alone cannot fully ensure cross-view consistency, the globally optimized 3D map from Gaussian Splatting (GS) inherently maintains consistency across views.

\vspace{-5pt}

\begin{table}[ht]
\centering
\begin{footnotesize}
\caption{\textbf{Impact of Color-Language Disentanglement (disent.) versus Joint Multi-Channel Optimization (joint).}}
\label{tab:disentangledGS:summary}
\setlength{\tabcolsep}{1pt}  
\begin{tabular}{l|ll|ll|lll|c}
\hline
\rowcolor{gray!10}
\multirow{1}{*}{\textbf{Method}} & \multicolumn{2}{c|}{\textbf{2D Loc.}} & \multicolumn{2}{c|}{\textbf{3D Loc.}} & \multicolumn{3}{c|}{\textbf{Image Rendering}} & \multicolumn{1}{c}{\textbf{Tracking}} \\
\cline{2-9}
\rowcolor{gray!10}
 & mIOU$\uparrow$ & Loc$\uparrow$ & EMD$\downarrow$ & CD $\downarrow$ & PSNR$\uparrow$ & SSIM$\uparrow$ & LPIPS$\downarrow$ & ATE[cm]$\downarrow$\\
\hline
Joint   & 0.323 & \textbf{0.633} & 0.384 & 1.940 & 31.23 & 0.901 & 0.197 & 0.796\\
Disent.  & \textbf{0.402} & 0.622 & \textbf{0.375} & \textbf{0.974} &\textbf{35.89} & \textbf{0.957} & \textbf{0.060} & \textbf{0.325} \\
\hline
\end{tabular}
\end{footnotesize}
\end{table}
\noindent{\textbf{Online Learning of AutoEncoding (OLAE).}} 
The ablation study on Online Learning of AutoEncoder (OLAE) in CLIP Compression is summarized in Table~\ref{tab:replica_results:summary} and supplemental Table~\ref{tab:replica_results}. To evaluate the effect of removing the online encoder strategy, we train a single autoencoder using 4-fold cross-validation on the Replica dataset. In each fold, two sequences are held out for testing, while the remaining sequences are used for training.
This setup ensures that the trained modules are exposed to data from the Replica domain while making that the testing scenes remain unseen.

As shown in Table~\ref{tab:replica_results:summary}, introducing OLAE even surpasses the in-domain fine-tuned single autoencoder, demonstrating its effectiveness in preserving semantic concepts upon compression. A more detailed per-scene analysis (supplemental Table~\ref{tab:replica_results}) further reveals: Although in-domain fine-tuned pipelines tend to outperform on certain testing scenes similar to those observed during training, OLAE performs better on novel scenes and shows higher overall stability across all scenes. 
A more detailed ablation study on code size is provided in supplemental Sec.~\ref{supp:code:size}.

\noindent{\textbf{Color-Language Disentanglement.}}
We evaluate the impact of disentanglement by comparing it to joint multi-channel optimization commonly used in other online multi-modality learning frameworks~\cite{li2024gs3lam,ji2024neds,li2025sgs}. This study uses the Replica Room-0 subset to examine the design's impact on 2D / 3D localization accuracy, novel-view image rendering quality and SLAM tracking errors (ATE [cm]), with results shown in Table~\ref{tab:disentangledGS:summary}.


The results demonstrate that disentangling color and language significantly enhances 2D mIoU, both 3D metrics, color rendering, and camera tracking, while maintaining comparable 2D localization performance. A more detailed ablation study on the impact of GS parameters is provided in supplemental Sec~\ref{supp:disentangle}, confirming that this strategy enables the two modalities to operate with their optimal GS parameters, minimizing interference between them. 


\vspace{-3pt}
\section{Conclusion}
\vspace{-3pt}
\label{sec:conclusion}
In this work, we introduce Online Language Splatting, a framework that enables online language-aware 3D mapping through key innovations. First, a real-time Super-Resolution Decoder (SRD) enhances CLIP embeddings, generating detailed language maps. Second, an highly effective and efficient two-stage CLIP compression preserving open-vocabulary capabilities. Third, a color-language disentangled optimization improves rendering quality for both language and color images. 
Our experimental results demonstrate that our online approach not only outperforms offline SoTA Lang-GS methods, but also leads to orders of magnitude efficiency improvement.


{
    \small
    \bibliographystyle{ieeenat_fullname}
    \bibliography{main}
}

\clearpage
\maketitlesupplementary

\section{Super-Resolution Decoder Architecture}
\label{supp:super-res}
\begin{figure}[bt!]
    \centering
    \includegraphics[trim=1mm 1.0mm 3.0mm 0.0mm,clip, width=1.0\columnwidth]{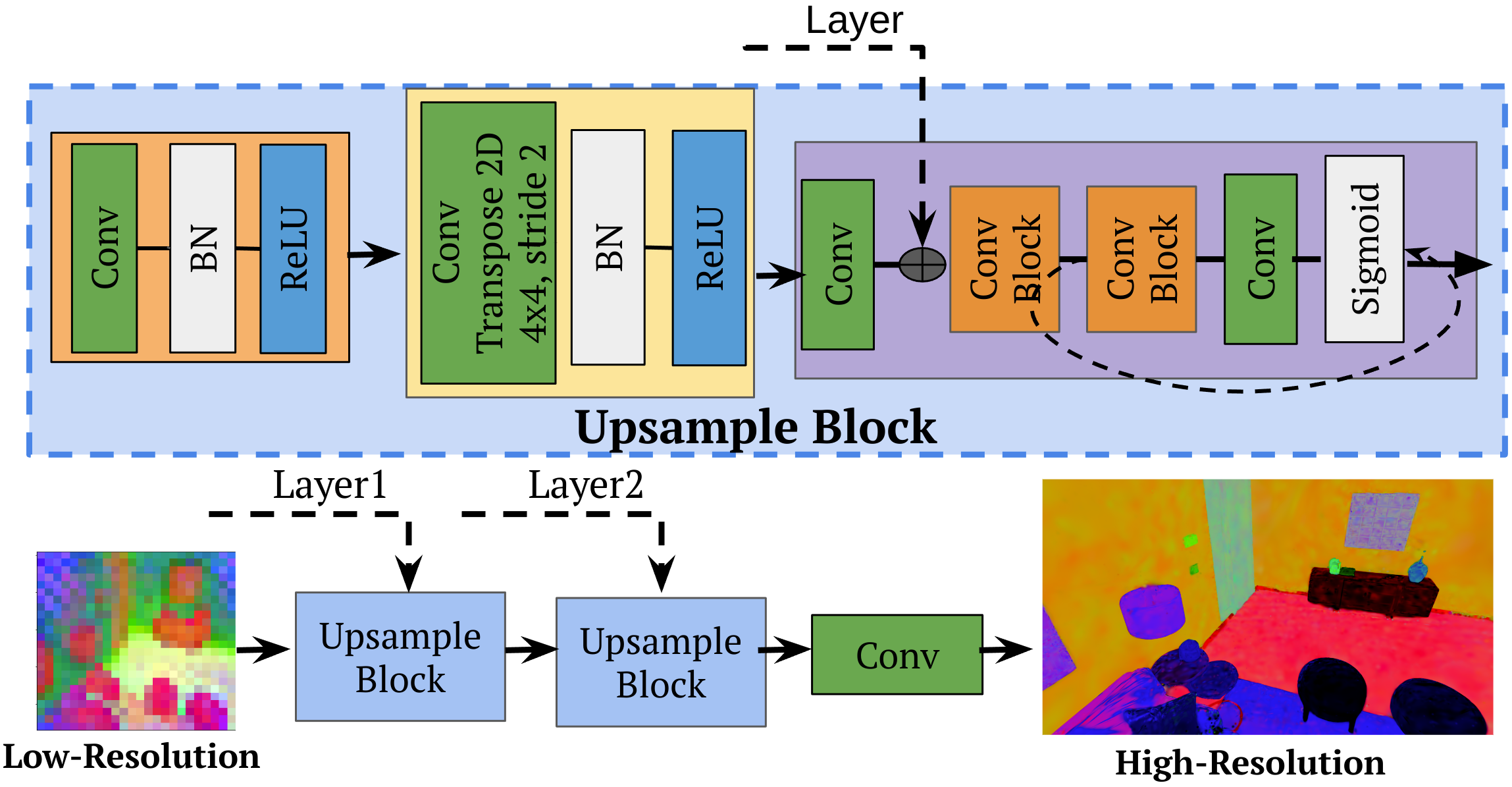}
    \vspace{-18pt}
    \caption{
    \textbf{Super-Resolution Decoder (SRD) Architecture.} The architecture consists of multiple layers designed to transform low-resolution input features into high-resolution outputs. The process begins with Upsample Blocks, each composed of convolutional layers, batch normalization (BN), ReLU activation, and ConvTranspose layers for spatial upscaling. After successive upsampling with the fusion of encoder intermediate layers' outputs, the output passes through a final convolutional block and sigmoid activation to produce the high-resolution feature map. This decoder refines low-resolution language features into detailed, pixel-aligned high-resolution maps for enhanced spatial understanding.
    }
    \label{fig:srd_network}
\end{figure}

We illustrate the detailed network architecture of our super-resolution decoder, as shown in Fig.~\ref{fig:srd_network}. Since the design only includes low-cost CNN layers, this module achieves real-time. Together with the pixel-wise CLIP encoder, they compose the real-time high-resolution CLIP embedding module adopted in our online language splatting framework.
For training, we use a combination of losses to ensure high-resolution feature quality and semantic coherence. The loss function is defined as:
\[
\mathcal{L} = 0.8 \cdot \mathcal{L}_{\text{cosine}} + \mathcal{L}_{\text{L1}} + 0.01 \cdot \mathcal{L}_{\text{TV}},
\]
\begin{itemize}
    \item \(\mathcal{L}_{\text{L1}}\) is the L1 loss, computed as:
    \[
    \mathcal{L}_{\text{L1}} = \frac{1}{N} \sum_{i=1}^{N} \left| \mathbf{y}_{\text{pred}} - \mathbf{y}_{\text{gt}} \right|,
    \]

    \item \(\mathcal{L}_{\text{cosine}}\) is the cosine similarity loss, computed as:
    \[
    \mathcal{L}_{\text{cosine}} = 1 - \frac{\mathbf{y}_{\text{pred}} \cdot \mathbf{y}_{\text{gt}}}{\|\mathbf{y}_{\text{pred}}\| \|\mathbf{y}_{\text{gt}}\|},
    \]

    \item \(\mathcal{L}_{\text{TV}}\) is the total variation loss, computed as:
    \[
    \mathcal{L}_{\text{TV}} = 
    \sum_{i, j} \left| \mathbf{y}_{\text{pred}}(i, j) - \mathbf{y}_{\text{pred}}(i, j+1) \right| 
    \]
    \[
    + \sum_{i, j} \left| \mathbf{y}_{\text{pred}}(i, j) - \mathbf{y}_{\text{pred}}(i+1, j) \right|.
    \]

    where \(\mathbf{y}_{\text{pred}}\) is the predicted high-resolution feature map, and \(\mathbf{y}_{\text{gt}}\) is the ground-truth feature map. The \(\mathcal{L}_{\text{TV}}\) loss penalizes spatial discontinuities to ensure smoothness \cite{fu2024featup}.
\end{itemize}

\textcolor{black}{
SRD is supervised using labels that are created from SAM-generated masks. For each image, multiple points are sampled, clustered, and refined to produce the most accurate mask, which is then propagated consistently across the labels. We train the model on the COCO and Omni datasets, leveraging hierarchical features and supervised labels to help the network learn to associate boundaries and propagate information effectively. This enables the network to produce high-quality, generalizable language features.
}

\section{Comparison to Language-GS SoTA methods on Replica Per Scene}

We present the complete comparison to Language-GS SoTA methods on each scene of Replica Dataset in Table~\ref{tab:replica_results}. As observed, our method achieve overall SoTA in both mIOU and LOC cross all scenes. On the ohter hand, certain inconsistency is also observed cross views. This may stem from varying domain gaps between testing scene and AE pretraining domains. The online AE, pretrained on the COCO and fine-tuned online, exhibits consistent results. In contrast, rows 1 and 2 (ours w/o online) use AEs trained on other Replica scenes, where greater divergence from the testing scene may cause inconsistency.

For evaluation, we utilize the following top 10 labels per scene:
\begin{table}[t]
\centering
\footnotesize
\caption{\textbf{Top 10 Labels Used for Evaluation in Each Scene of the Replica Dataset.}}
\vspace{-5pt}
\label{tab:replica_labels}
\setlength{\tabcolsep}{2pt}
\begin{tabular}{l|p{8cm}}
\hline
\rowcolor{gray!10}
\textbf{Scene} & \textbf{Top 10 Labels} \\
\hline
Office0  & wall, rug, table, blinds, sofa, tv-screen, chair, floor, door, bin \\
Office1  & wall, floor, pillow, blanket, blinds, desk, desk-organizer, monitor, table, chair \\
Office2  & wall, floor, table, sofa, panel, cushion, chair, tv-screen, bottle, tissue-paper \\
Office3  & floor, table, wall, window, chair, sofa, tablet, cushion, door, switch \\
Office4  & wall, floor, chair, ceiling, window, bench, panel, tv-screen, table, clock \\
Room0    & wall, window, floor, sofa, cushion, table, rug, lamp, book, indoor-plant \\
Room1    & wall, window, blinds, floor, blanket, lamp, ceiling, comforter, nightstand, picture \\
Room2    & wall, chair, floor, plate, vase, window, table, indoor-plant, rug, shelf \\
\hline
\end{tabular}
\vspace{-2pt}
\end{table}

\input{sec/4_1_replica_table}

\begin{figure}
    \centering
    \includegraphics[width=1.0\linewidth]{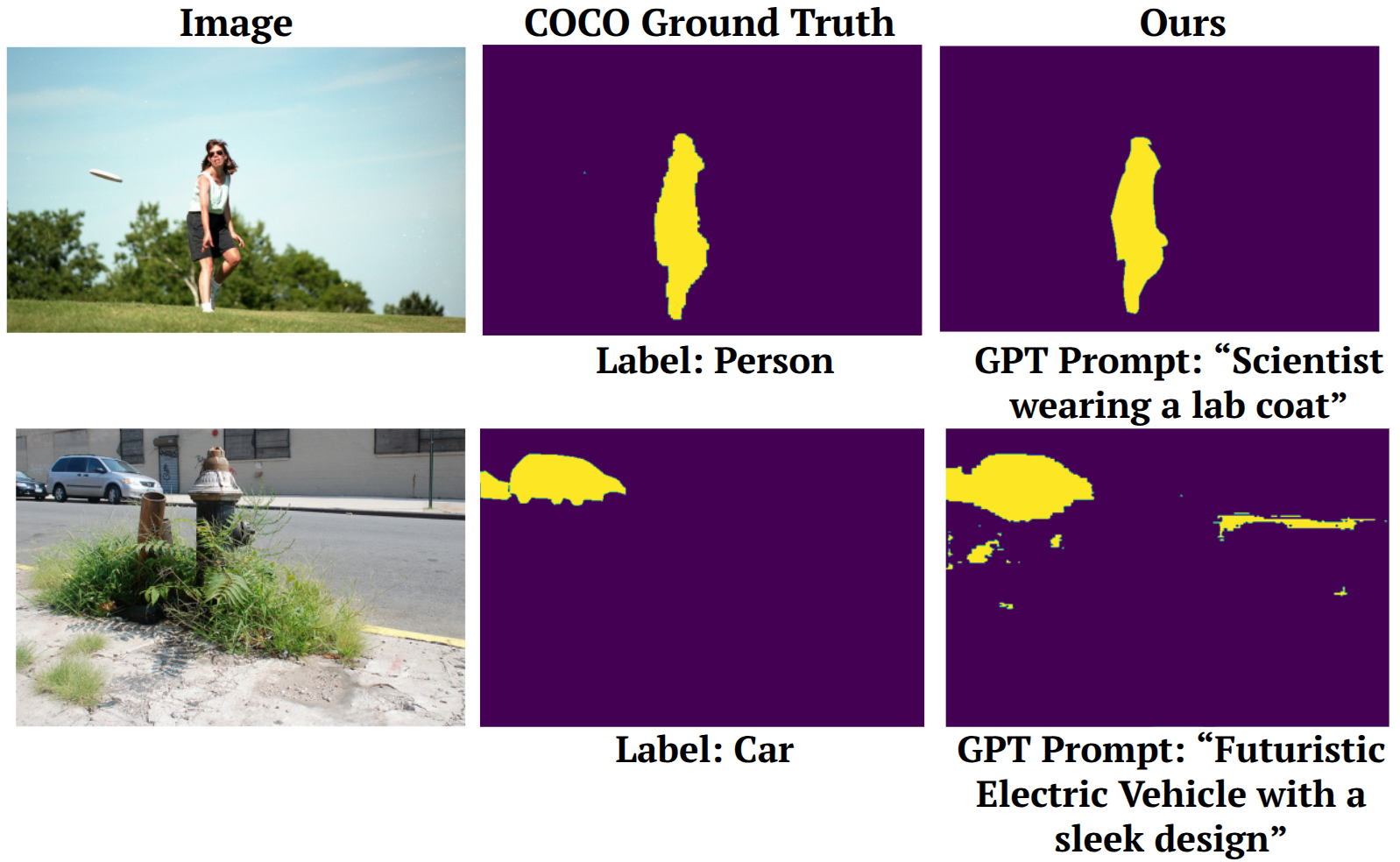}
    \includegraphics[width=1.0\linewidth]{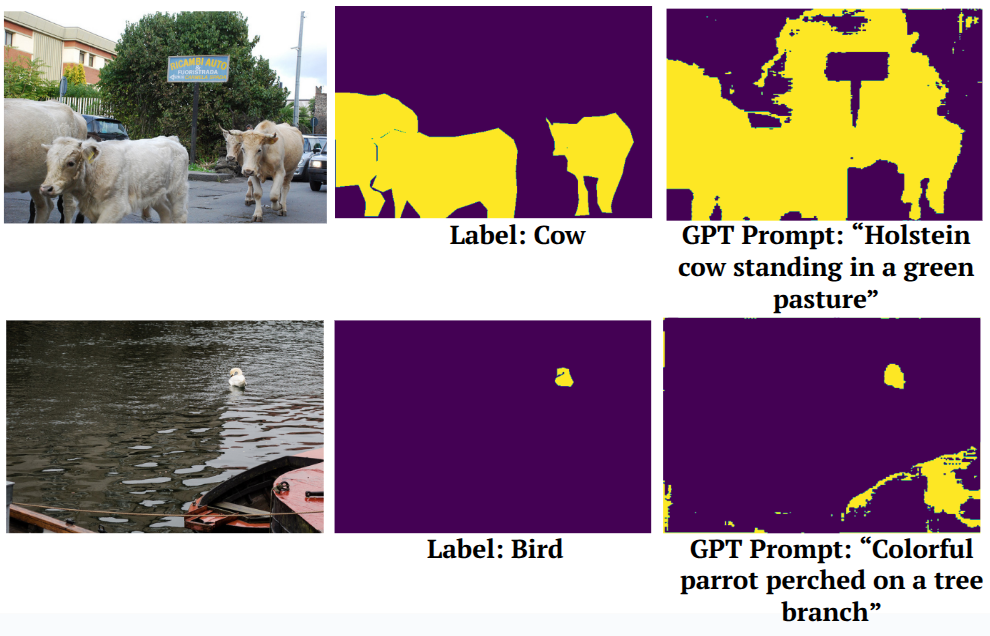}
    \caption{Open-vocabulary segmentation. \textbf{Left}: COCO ground truth segmentation. \textbf{Right}: Segmentation output of our module using GPT-generated novel vocabulary prompt.}
    \label{fig:gpt_label_ov}
\end{figure}

\section{Comparison to SLAM-GS SoTA methods on Replica Per Scene}

We present the complete SLAM-GS evaluation results for each scene in Table~\ref{tab:SOTA_comparison_slam_gs}. As observed, although our method incorporates additional open-vocabulary language mapping, it maintains the novel view rendering quality of the baseline MonoGS. Overall, our approach achieves state-of-the-art (SoTA) performance in PSNR and LPIPS metrics.
\input{sec/4_slam_sota_table}

\section{Open-Vocabulary Evaluation}
We compare our method with other offline Lang-GS methods using GPT-generated labels on Replica. We use GPT-4o with the following prompt:
"Describe the image with 5 vocabularies for each image to test object segmentation."
We randomly select 30 images from 8 Replica sequences and generate labels, which are then used as prompts to query objects for these images. To generate segmentation masks and ground truth, we use Grounded SAM~\cite{ren2024grounded}, leveraging the GPT-selected open-vocabulary (OV) labels as queries.

The evaluation results, comparing our method against offline SoTA Lang-GS methods, are presented in Table~\ref{tab:replica_results:ov}. While our method ranks secondary to LangSplat, it still outperforms Feature3DGS and LEGaussian by large margins and operates over $40\times$ faster, highlighting its efficiency and strong open-vocabulary segmentation performance. 

\begin{figure*}
    \centering
    \includegraphics[width=1.0\linewidth]{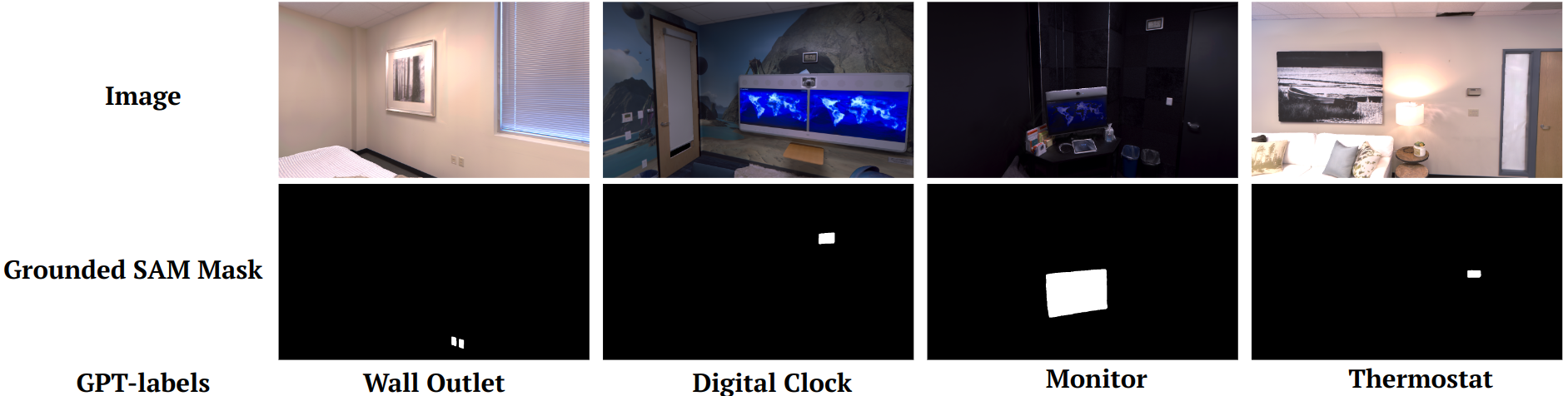}
    \caption{Examples of GPT-generated object labels and masks from Grounded SAM.}
    \label{fig:gpt_label_example}
\end{figure*}

We found that the key factors determining generalization capability to open-vocabulary (OV) objects is the resolution of feature maps used for GS mapping.
As observed, the GPT-generated labels include many tiny objects such as “thermostat”, “wall outlet”, and “digital clock”, which are difficult to detect in low-resolution feature maps (See Fig.~\ref{fig:gpt_label_example} for examples.) 
Our method operates at a spatial resolution (192×192) using SRD, which is much higher compared to the pixel-wise encoder output (32x32), but remains constrained by the speed requirement for online integration of language features into 3DGS.  This resolution may pose challenges for detecting tiny objects, however it provides a significant advantage in running speed, making our approach suitable for online SLAM applications.
In contrast, LangSplat operates at full-resolution feature maps (1200×680), embedding them directly into 3DGS, which enhances tiny object detection but comes at the cost of a much slower runtime, making it unsuitable for real-time SLAM applications.

We acknowledge tiny object detection as a limitation of our current approach and discuss it further as part of our future work in Sec.~\ref{sec:limitation}.

Additionally, we evaluate the open-vocabulary segmentation of our model to determine whether it preserves the ability to segment objects using novel textual descriptions as prompts beyond the original COCO vocabulary. To test this, we randomly sample 100 COCO test images and use ChatGPT (GPT-4o) to generate semantically richer descriptions for each label, such as replacing "car" with "futuristic electric vehicle with a sleek design". Using our trained model, which leverages CLIP-based feature representations, we generate segmentation masks for both the COCO labels and the GPT-generated descriptions. The model achieves an mIoU of 0.389 with COCO labels and 0.392 with GPT labels (Table~\ref{tab:miou_comparison}, demonstrating that it maintains segmentation performance regardless of textual variation. The qualitative results (Fig.~\ref{fig:gpt_label_ov} compare the COCO ground-truth segmentation with our model's segmentation using GPT-generated novel descriptions. The results support the ability to generalize beyond COCO labels. These findings confirm that despite being trained on COCO for upsampling, the model effectively operates in an open-vocabulary setting. 

\begin{table}[h]
    \caption{Comparison of mIoU performance using COCO dataset labels and ChatGPT-generated novel vocabulary on 100 randomly sampled test images. The similar mIoU scores indicate that our method preserves CLIP's open-vocabulary capabilities.}
    \centering
    \begin{tabular}{l c}
        \toprule
        Method & mIOU \\
        \midrule
        COCO labels & 0.389 \\
        GPT novel labels & 0.392 \\
        \bottomrule
    \end{tabular}
    \label{tab:miou_comparison}
\end{table}

\begin{table}[t]
\centering
\footnotesize
\caption{\textbf{Comparison on GPT-generated labels for Replica}. [Key: \colorbox{brightred}{best}, \colorbox{lightred}{second-best}] }
\vspace{-5pt}
\label{tab:replica_results:ov}
\setlength{\tabcolsep}{3pt}
\begin{tabular}{l|cc|c|c}
\hline
\rowcolor{gray!10}
{\textbf{Method}} & \multicolumn{2}{c|}{\textbf{GPT-labels}} & 
{\textbf{FeatureMap Res.}} &{\textbf{Time}} \\
\cline{2-3}
\rowcolor{gray!10}
 & mIOU & Loc & \\
\hline
LangSplat~\cite{qin2024langsplat} & \colorbox{brightred}{0.660} & \colorbox{brightred}{0.880} & 1200$\times$680 & \colorbox{lightred}{2.8 min/fr} \\
Feature3DGS~\cite{zhou2024feature3dgs} & 0.489 & 0.600 & 480$\times$360 & 2.3 min/fr\\
LEGaussian~\cite{shi2024language} & 0.241 & 0.703 & 184$\times$110 &32 s/fr \\
Ours & \colorbox{lightred}{0.539} & \colorbox{lightred}{0.765} & 192$\times$192 &\colorbox{brightred}{0.8 s/fr} \\ 
\hline
\end{tabular}

\vspace{-2pt}

\end{table}

\begin{figure}[bt!]
\centering
\includegraphics[trim=1mm 1.0mm 3.0mm 0.0mm,clip, width=1.0\columnwidth]{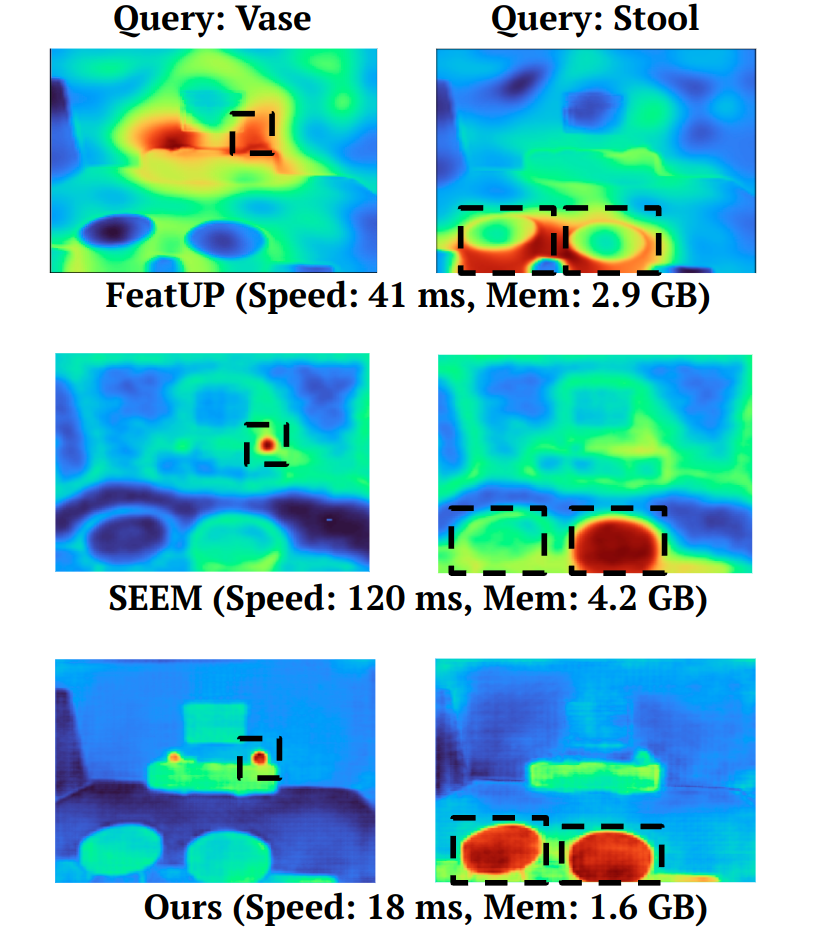}
\caption{
We provide a visual comparison for feature maps by FeatUP \cite{fu2024featup}, the open-vocabulary segmentor from SEEM \cite{zou2023segment}, and our HR module. As observed, despite its simplicity, our HR module achieves the highest feature quality at the lowest cost and fastest speed. We believe our simple and
highly effective design provides valuable new insights. Black box: groundtruth.
}
\label{fig:hr_compare}
\end{figure}

\section{More Visualizations}

\begin{figure*}[t]
    \centering
    \includegraphics[trim=5mm 0mm 0mm 0mm, clip, width=0.90\linewidth]{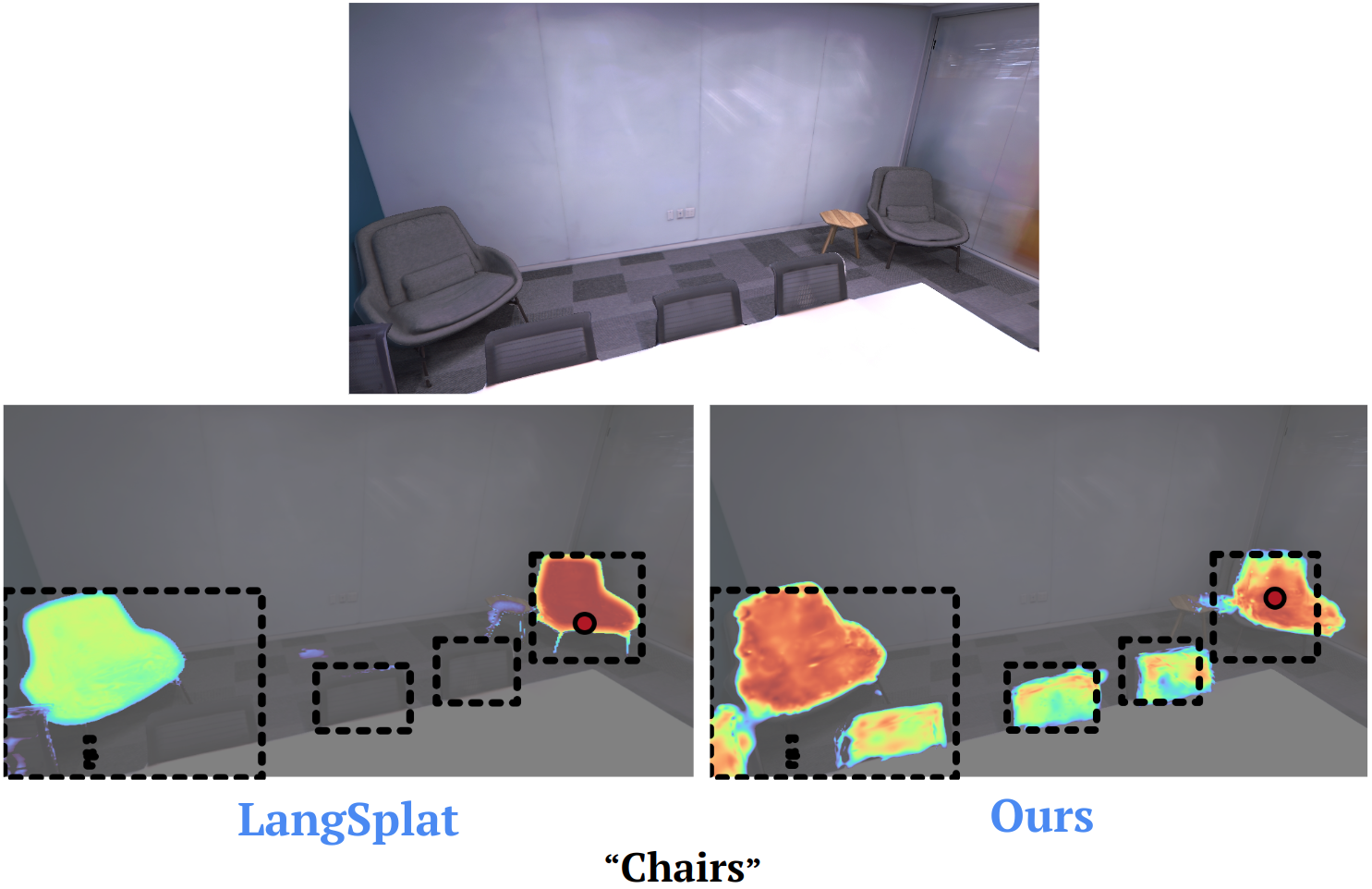}
    \caption{Our method is able to identify and segment all of the chairs, while LangSplat was only able to segment only two chairs.}
    \label{fig:supp_fig1}
    
    
    \includegraphics[trim=5mm 0mm 0mm 0mm, clip, width=0.90\linewidth]{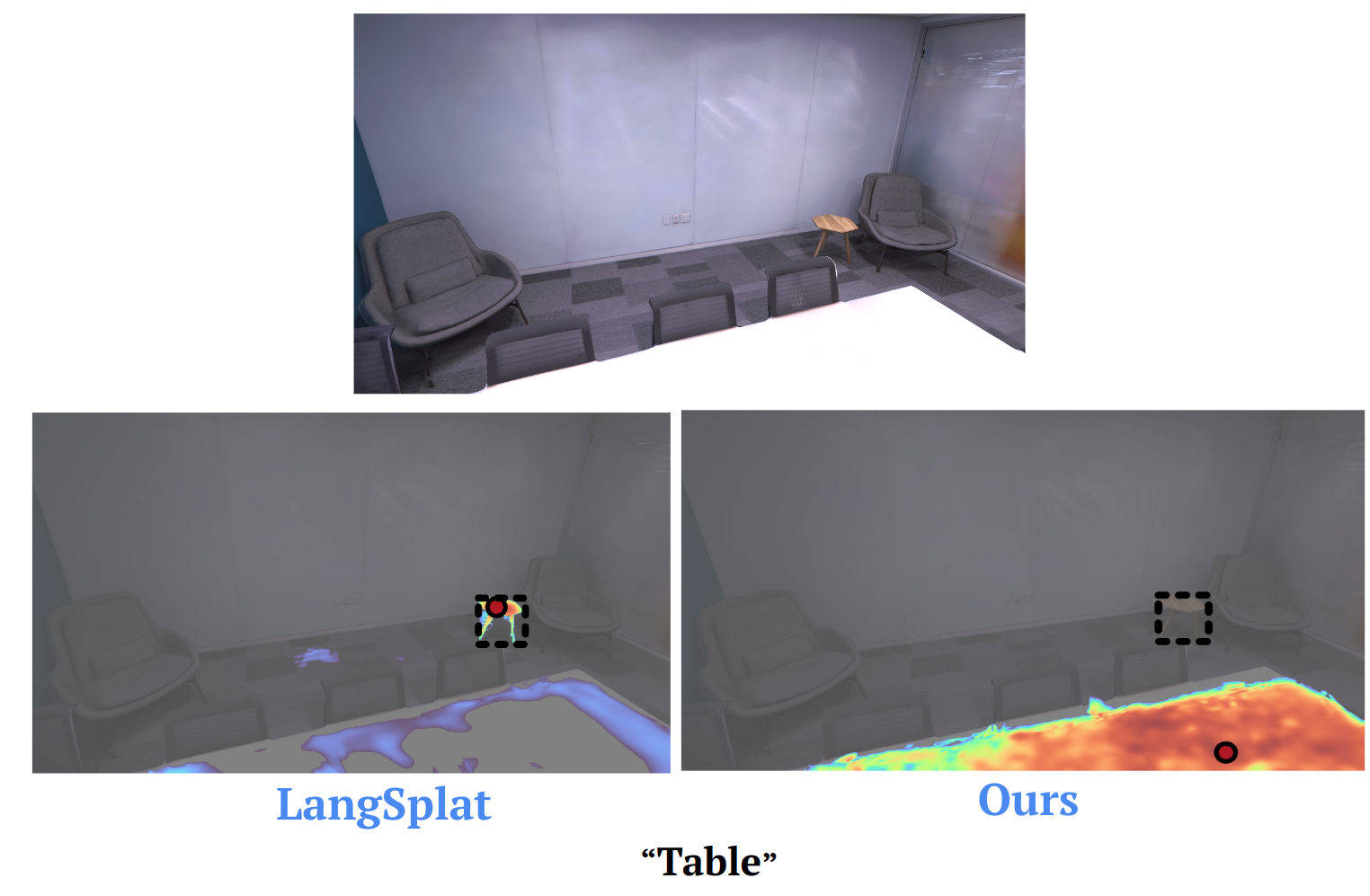}
    \caption{Comparison of segmentation results for the query "Table". LangSplat successfully segments the table, while our method focuses on the table top instead, leading to a segmentation error. This discrepancy is considered a failure case for our approach.}
    \label{fig:supp_fig2}
\end{figure*}

\begin{figure*}[t]
    \centering
    \includegraphics[trim=5mm 0mm 0mm 0mm, clip, width=0.95\linewidth]{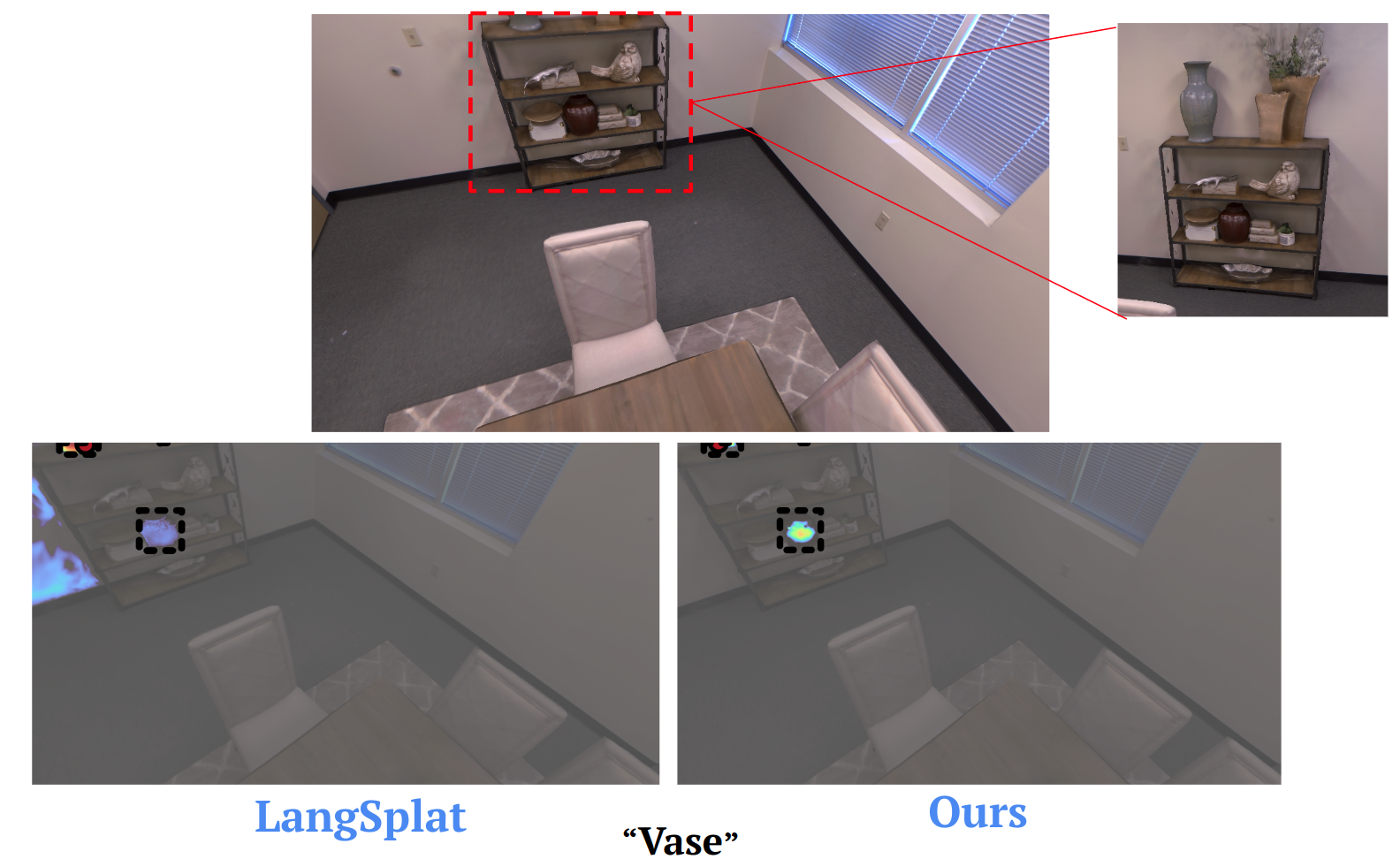}
    \caption{Both methods demonstrate 3D consistency. The top-right zoom-in image shows the vase from another frame. Our model, due to its 3D consistency, successfully detects the vase from only very limited appearance in the current frame. This highlights the ability of our method to handle occlusions or partial appearance effectively.}
    \label{fig:supp_fig3}
    
    \vspace{5mm} 
    
    \includegraphics[trim=5mm 0mm 0mm 0mm, clip, width=0.95\linewidth]{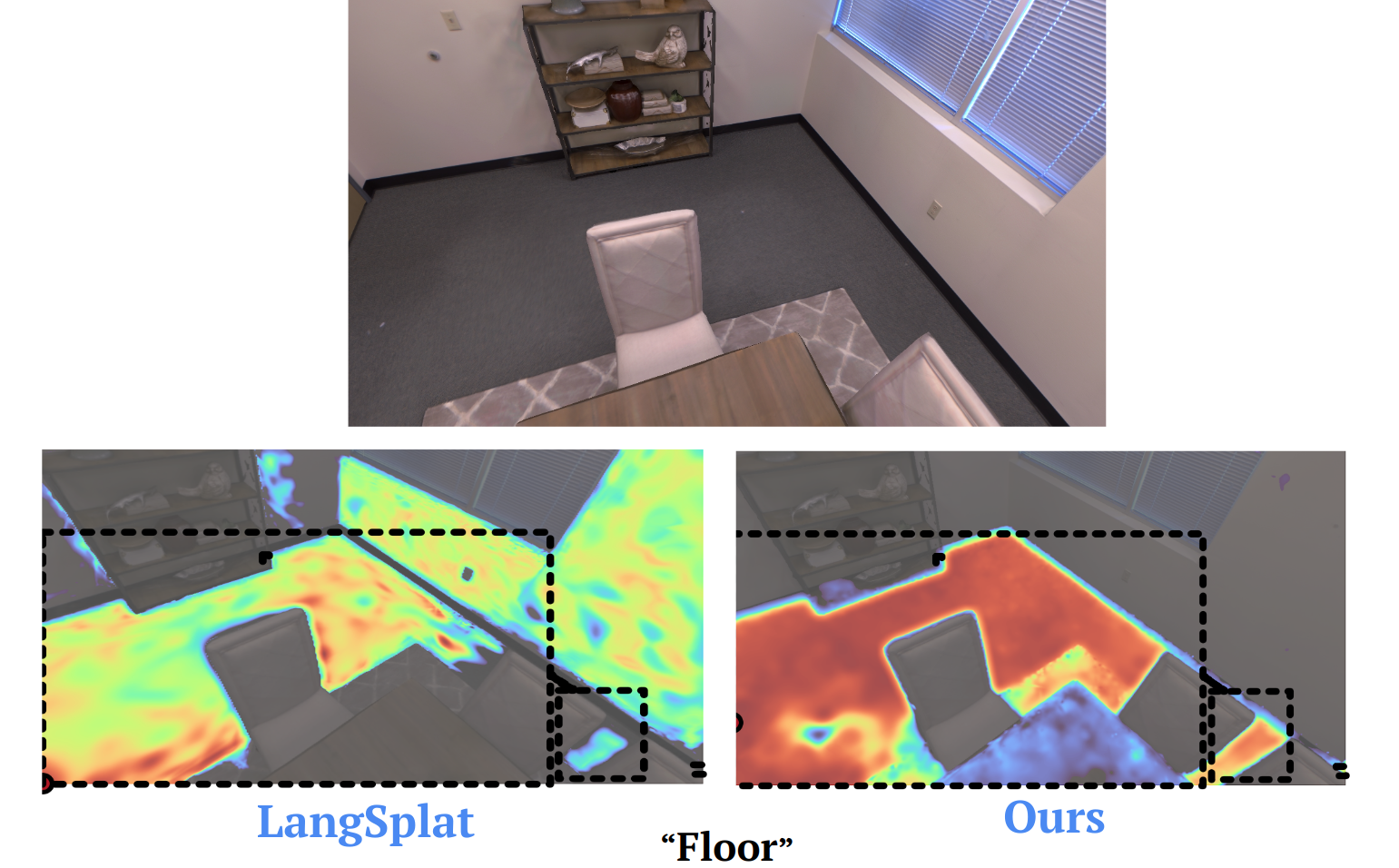}
    \caption{Comparison of floor segmentation results. LangSplat introduces outliers by incorrectly segmenting walls as part of the floor. In contrast, our method accurately segments the floor without including such outliers.}
    \label{fig:supp_fig4}
\end{figure*}

\begin{figure*}[t]
    \centering
    \includegraphics[trim=5mm 0mm 0mm 0mm, clip, width=0.95\linewidth]{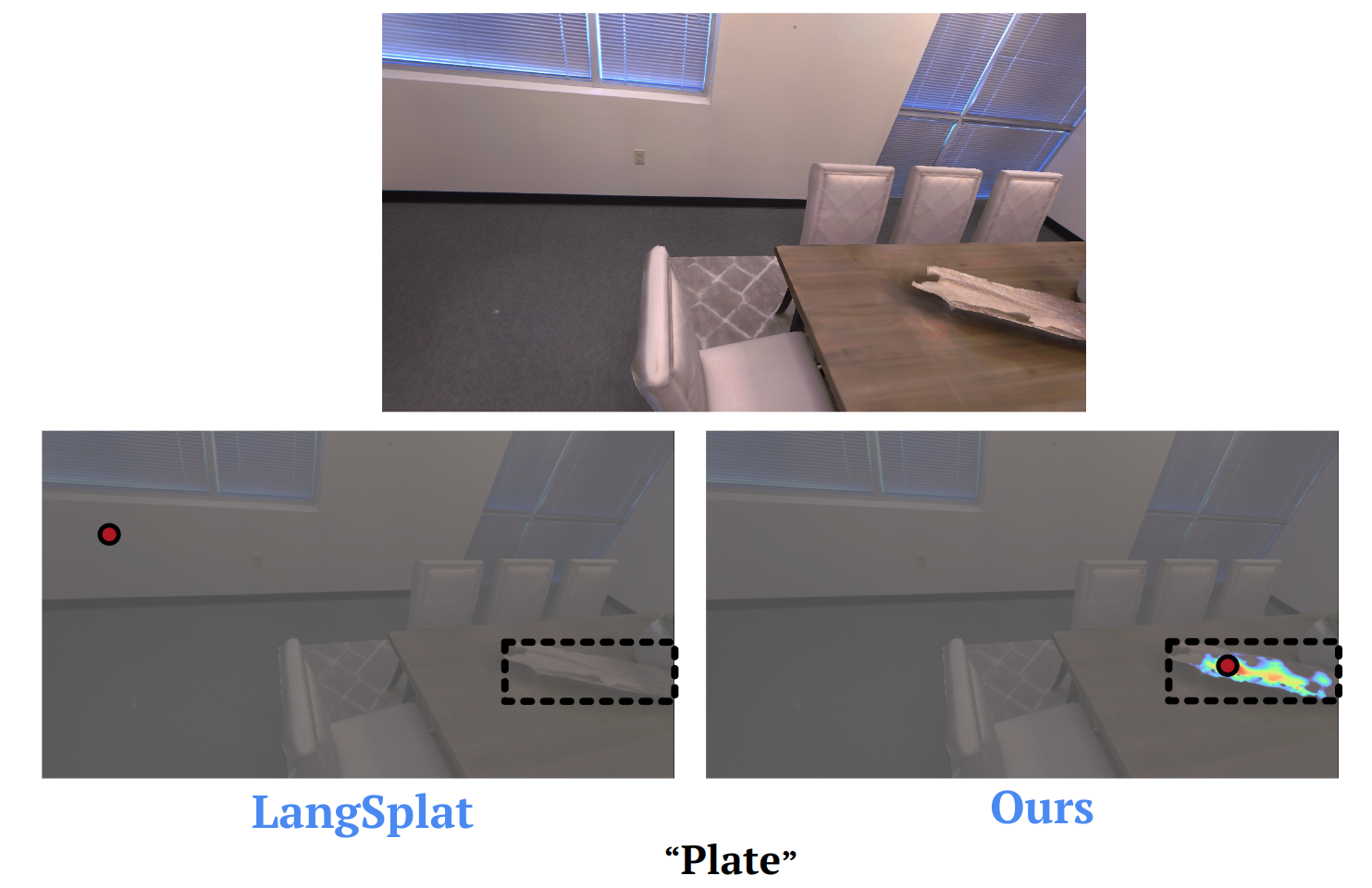}
    \label{fig:supp_fig5}
    \caption{Comparison of plate segmentation. LangSplat fails to detect the plate, whereas our method successfully identifies and segments the plate on the table.}
\end{figure*}

\begin{figure*}[t]
    \centering
    \includegraphics[trim=5mm 0mm 0mm 0mm, clip, width=0.75\linewidth]{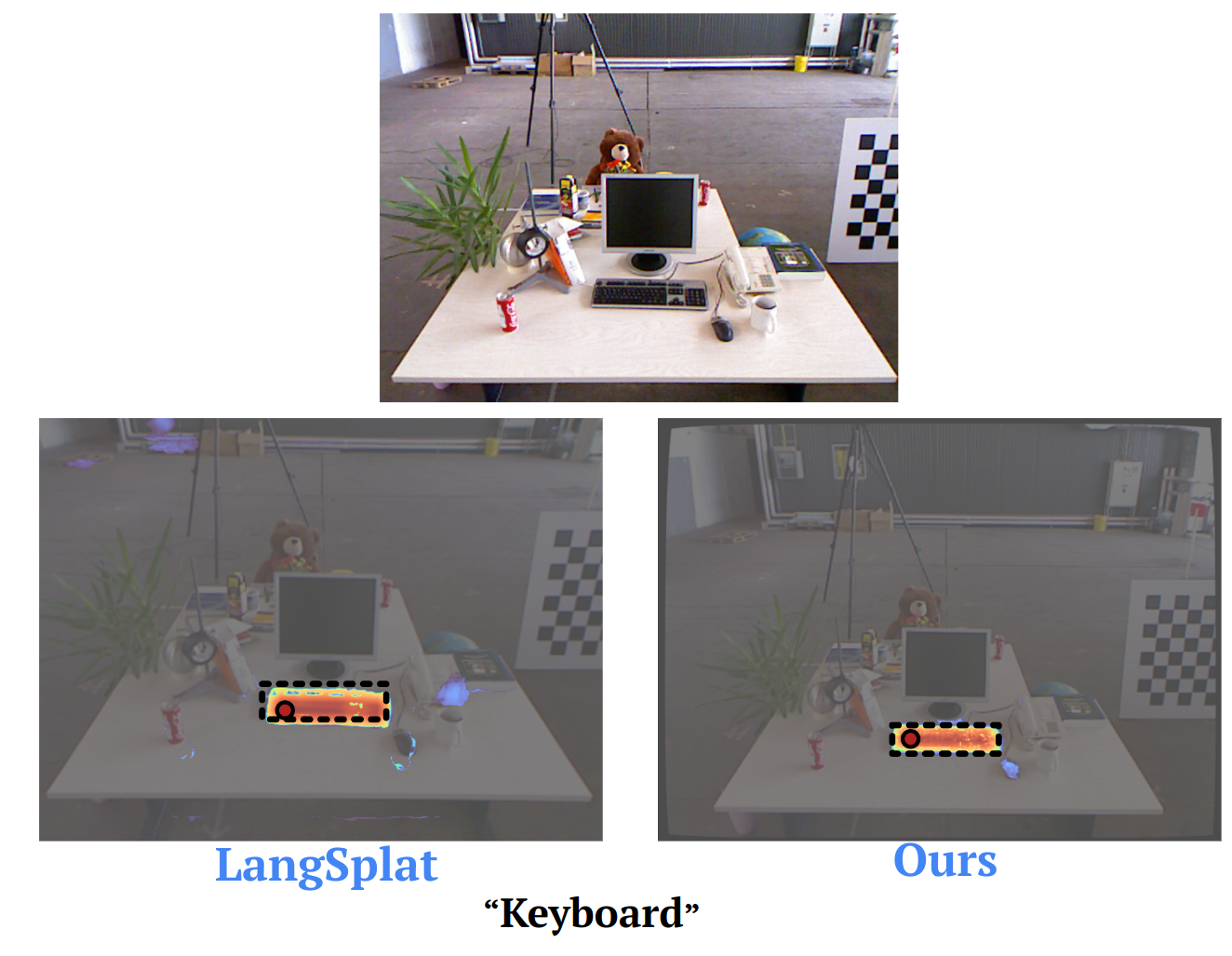}
    \caption{Comparison of "Keyboard" query localization on TUM-RGBD.}
    \label{fig:supp_fig6}
    
    \vspace{5mm} 
    
    \includegraphics[trim=5mm 0mm 0mm 0mm, clip, width=0.75\linewidth]{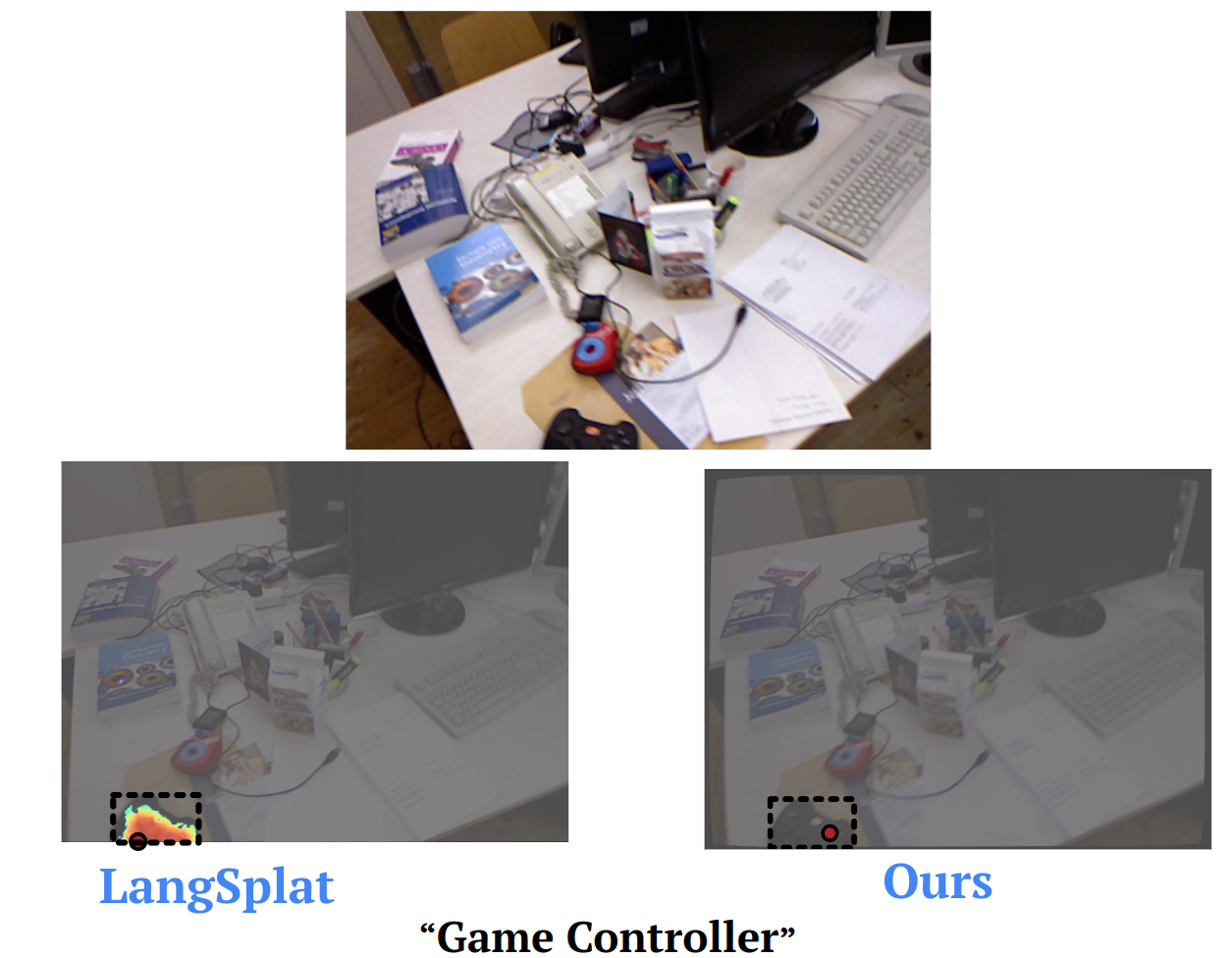}
    \caption{Comparison of "Game Controller" query localization on TUM-RGBD.}
    \label{fig:supp_fig7}
\end{figure*}

In this section, we provide additional visualization results in Figs.~\ref{fig:supp_fig1}-~\ref{fig:supp_fig7}. To ensure a fair comparison, we increase LangSplat's code size from 3 to 15 and upgrade OpenCLIP's \cite{ilharco_gabriel_2021_5143773} feature dimension from 512 to 768.


\textbf{Details of heat map results and evaluation metrics.} We display 2D heat maps as query results throughout this work. For each text query, LangSplat generates three Gaussian relevancy language features, while our method produces pixel-level language features through our high-resolution model. To calculate localization and IoU metrics and reduce the impact of outliers, similar to LangSplat, we apply a mean convolution filter with a kernel size of 20 to smooth the values in the language feature maps. The final score is determined by selecting the maximum relevancy score.

We used a score threshold of 0.4 for LangSplat and 0.5 for our method. We tuned the threshold that shows the best object boundaries for each method for a fair comparison. Values below the threshold are classified as background, and those above it generate binary maps.


\begin{table}[ht]
\centering
\footnotesize
\caption{Comparison of various language code sizes on Replica Office0 (Of0) and Room2 (Rm2) sequences.}
\label{tab:coco_ae}
\setlength{\tabcolsep}{4pt}  
\begin{tabular}{cccccc}

\toprule
 Size & & \textbf{128} & \textbf{64} & \textbf{32} & \textbf{20} \\
\midrule
\multirow{3}{*}{Rm2} 
& MSE & 0.0064 & 0.0065 & 0.0065 & 0.0068 \\
& MAE & 0.0497 & 0.0502 & 0.0505 & 0.0620 \\
& Cosine Sim. & 0.9817 & 0.9764 & 0.9690 & 0.9000 \\
\midrule
\multirow{3}{*}{Of0} 
& MSE & 0.0059 & 0.0060 & 0.0068 & 0.0074 \\
& MAE & 0.0439 & 0.0445 & 0.0484 & 0.0570 \\
& Cosine Sim. & 0.9823 & 0.9750 & 0.8790 & 0.7600 \\
\bottomrule
\end{tabular}
\end{table}

\begin{table}[ht]
\centering
\footnotesize
\caption{Comparison of IOU and Localization (Loc) accuracy across different code sizes (15, 6, 3) for Replica Office3 (Of3) and Office4 (Of4) sequences.}
\label{tab:online_ae}
\setlength{\tabcolsep}{4pt}  
\begin{tabular}{lcccc}
\toprule
\textbf{Code Size} & \textbf{Of3 IOU} & \textbf{Of3 Loc} & \textbf{Of4 IOU} & \textbf{Of4 Loc} \\
\midrule
15 & 0.495 & 0.766 & 0.498 & 0.765 \\
6 & 0.485 & 0.708 & 0.490 & 0.701 \\
3 & 0.480 & 0.690 & 0.487 & 0.693 \\
\bottomrule
\end{tabular}
\end{table}

\begin{table*}[h!]
\centering
\footnotesize
\caption{\textbf{Comparison on 3D localization evaluation}. We counts a query as failure when a distance is larger than the CD/EMD's population mean plus 2$\times$ population standard deviation. Failures are excluded from the average and reported separately. Of: Office; Rm: Room from the Replica Dataset.}
\label{tab:3d_eval_all}
\setlength\tabcolsep{1pt}
\begin{tabular}{lcc|cc|cc|cc|cc|cc|cc|cc|cc|}
\toprule
Average CD & \multicolumn{2}{c|}{\textbf{Of0}} & \multicolumn{2}{c|}{\textbf{Of1}} & \multicolumn{2}{c|}{\textbf{Of2}} &\multicolumn{2}{c|}{\textbf{Of3}}  & \multicolumn{2}{c|}{\textbf{Of4}}  & \multicolumn{2}{c|}{\textbf{Rm0}}  & \multicolumn{2}{c|}{\textbf{Rm1}}  & \multicolumn{2}{c|}{\textbf{Rm2}} & \multicolumn{2}{c|}{\textbf{Overall}}\\
& CD & Failure & CD & Failure & CD & Failure & CD & Failure & CD & Failure & CD & Failure & CD & Failure & CD & Failure & CD & Total count\\
\midrule
LangSplat~\cite{qin2024langsplat} & 1.175 & 1 & 0.764 & 3 & 1.232 & 1 & 0.828 & 1 & 1.450 & 1 & 0.942 & 1 & 1.044 & 1 & 0.342 & 1 & \cellcolor{thirdorange}0.972& \cellcolor{thirdorange}\textbf{10}\\
Ours & 0.620 & 1 & 0.922 & 3 & 0.804 & 1 & 0.284 & 1 & 1.380 & 0 & 0.657 & 1 & 0.582 & 2 & 0.567 & 1 & \cellcolor{thirdorange}\textbf{0.736}& \cellcolor{thirdorange}\textbf{10}\\
\bottomrule
\end{tabular}

\begin{tabular}{lcc|cc|cc|cc|cc|cc|cc|cc|cc|}
\toprule
Average EMD & \multicolumn{2}{c|}{\textbf{Of0}} & \multicolumn{2}{c|}{\textbf{Of1}} & \multicolumn{2}{c|}{\textbf{Of2}} &\multicolumn{2}{c|}{\textbf{Of3}}  & \multicolumn{2}{c|}{\textbf{Of4}}  & \multicolumn{2}{c|}{\textbf{Rm0}}  & \multicolumn{2}{c|}{\textbf{Rm1}}  & \multicolumn{2}{c|}{\textbf{Rm2}} & \multicolumn{2}{c|}{\textbf{Overall}}\\
& EMD & Failure & EMD & Failure & EMD & Failure & EMD & Failure & EMD & Failure & EMD & Failure & EMD & Failure & EMD & Failure & EMD & Total count\\
\midrule
LangSplat~\cite{qin2024langsplat} & 7.369 & 2 & 4.745 & 4 & 2.949 & 1 & 5.512 & 1 & 17.42 & 2 & 4.549 & 1 & 3.824 & 2 & 4.109 & 2 & \cellcolor{thirdorange}6.310 & \cellcolor{thirdorange}15\\
Ours & 1.574 & 1 & 2.292 & 5 & 9.001 & 1 & 7.949 & 1 & 9.157 & 1 & 1.498 & 1 & 13.987 & 1 & 0.100 & 3 & \cellcolor{thirdorange}\textbf{5.695}& \cellcolor{thirdorange}\textbf{14}\\
\bottomrule
\end{tabular}
\end{table*}

\begin{table*}[h!]
\centering
\caption{\textbf{Comparison of 1-stage and 2-stage methods}}
\label{tab:compressionStages}
\begin{tabular}{lcc}
\toprule
\textbf{Scene} & \textbf{1-stage (768 $\rightarrow$ 15) Offline} & \textbf{2-stage (768 $\rightarrow$ 32 Pretrained, 32 $\rightarrow$ 15 Online)} \\
\midrule
Room0   & 0.514,\;0.835 & 0.552,\;0.810 \\
Room1   & 0.427,\;0.839 & 0.505,\;0.939 \\
Room2   & 0.396,\;0.801 & 0.493,\;0.824 \\
Office0 & 0.422,\;0.761 & 0.433,\;0.774 \\
Office1 & 0.409,\;0.802 & 0.522,\;0.826 \\
\midrule
\textbf{Mean} & 0.434,\;0.808 & \textbf{0.501},\;\textbf{0.835} \\
\bottomrule
\end{tabular}
\end{table*}

\section{Language Compression}
\subsection{Study on Code Size}
\label{supp:code:size}

We study the generalizability of autoencoder code sizes trained on COCO and tested on Replica sequences (\textit{Room2 (Rm2)} and \textit{Office0 (Of0)}) (see Table~\ref{tab:coco_ae}), evaluating reconstruction quality using Mean Squared Error (MSE), Mean Absolute Error (MAE), and Cosine Similarity. MSE measures the squared differences between original and reconstructed features, MAE quantifies the absolute deviation, and Cosine Similarity assesses the alignment of feature vectors. As the code size decreases, we observe an increase in errors, which is expected due to the trade-off between compactness and information retention. Smaller code sizes create more compact feature representations but often reduce structural detail and granularity in the reconstructed language feature maps, leading to losses in semantic and spatial accuracy. This happens because smaller latent spaces constrain the feature encoding, causing a loss of fine-grained variations that are essential for precise language-based localization. To balance accuracy and computational efficiency, we choose code 32, which retains sufficient semantic fidelity while remaining efficient for real-time applications.

We evaluate the impact of varying the code size of the online encoder-decoder on IOU and Localization (Loc) metrics using Replica sequences. See Table~\ref{tab:online_ae}. To balance memory and ensure real-time feasibility, we limit the maximum code size to 15. Online training improves adaptability by fine-tuning representations for specific sequences, but its effectiveness depends on the code size. Smaller code sizes constrain the latent space, leading to a loss of fine-grained details and limiting the benefits of online training. In contrast, a code size of 15 strikes a balance between compression and capacity, allowing the model to leverage online adaptability while preserving semantic and spatial accuracy.

\begin{table}[h!]
\centering
\footnotesize
\caption{\textbf{Effects of disentangling GS parameters into color and language modes.} }
\vspace{-5pt}
\label{tab:study_disent}
\setlength{\tabcolsep}{4pt}  
\begin{tabular}{|p{1.3cm}p{1.3cm}p{1.3cm}|p{0.80cm}p{0.80cm}p{0.80cm}|c}
\hline
\multicolumn{3}{|c|}{\textbf{Disentangled Mode}} & \multicolumn{3}{c|}{\textbf{Image Rendering}}  \\
\cline{1-6}
 Separate $\alpha$ & Separate $\bm R$ & Separate $\bm S$ & PSNR$\uparrow$ & SSIM$\uparrow$ & LPIPS$\downarrow$\\
\hline
 
 \centering\xmark &  \centering\xmark &  \centering\xmark & 31.23 & 0.901 & 0.197 \\
  \centering\xmark &  \centering\cmark &  \centering\cmark & 31.79 & 0.915 & 0.177 \\
 \centering\cmark & \centering\xmark & \centering\xmark & 31.75 & 0.918 & 0.180 \\
 \centering\cmark & \centering\cmark & \centering\xmark & 32.80 & 0.929 & 0.146 \\
 \centering\cmark & \centering\xmark & \centering\cmark & 33.57 & 0.939 & 0.118 \\
 \centering\cmark & \centering\cmark & \centering\cmark & \textbf{35.89} & \textbf{0.957} & \textbf{0.060} \\
\hline
\end{tabular}
\vspace{-2mm}

\end{table}

\textcolor{black}{\subsection{Online Compression} Table ~\ref{tab:compressionStages} presents a comparison between single-stage and two-stage compression methods across multiple scenes. The results indicate that the two-stage method generally offers  improvements in both mIOU and localization accuracy, with certain scenes like Room1 demonstrating more noticeable gains (mIOU from 0.427 to 0.505 and localization accuracy from 0.839 to 0.939). 
While the single-stage method is simpler and more memory-efficient, it may lose important language features due to aggressive compression (768D directly to 15D). Conversely, the two-stage method introduces an intermediate compression stage, preserving these language features but at the cost of increased complexity. Therefore, the choice between these two methods ultimately depends on the user's specific requirements and constraints regarding accuracy versus resource utilization.}

\section{Detailed Study on Disentanglement GS Parameters}
\label{supp:disentangle}

First, as a more detailed version of the 3D localization evaluation compared to Table~\ref{tab:disentangledGS:summary}, we present per-class evaluation results in Table~\ref{tab:disentangledGS}. As observed, the disentangled optimization leads to better overall performance and significant improvements on some classes.

Next, we study different strategies in GS parameter disentanglement, including separating $\alpha$, $\bm R$ and $\bm S$ into color and language modes. In the forward rendering, we splat 3D Gaussians onto the 2D space and conduct alpha compositions using respective mode parameters to render color and languages.

In the back-propagation, if $\alpha$ is disentangled into color ($c$) and language ($f$) modes, we then calculate gradients by
\begin{equation}
\footnotesize
\vspace{-2pt}
      \frac{\partial \mathcal{L}}{\partial \alpha^c_i} = \frac{\partial \mathcal{L}}{\partial C} \frac{\partial C} {\partial \alpha^c_i} + \frac{\partial \mathcal{L}}{\partial D} \frac{\partial D}{\partial \alpha^c_i}, \quad      \frac{\partial \mathcal{L}}{\partial \alpha^f_i} = \frac{\partial \mathcal{L}}{\partial F} \frac{\partial F}{\partial \alpha_i^f},
\vspace{-1pt}
\label{eq:backprop}
\end{equation}
where $\frac{\partial \mathcal{L}}{\partial \alpha^c_i}$ and $\frac{\partial \mathcal{L}}{\partial \alpha^f_i}$ further propagates to $\bm R_i$ and $\bm S_i$ via the world-coordinate 3D covariance matrix $\bm \Sigma^{c/f}_i$. If $\alpha$ is not disentangled, the gradient terms in Eq.~\eqref{eq:backprop} are added together. 

The same rule applies to $\bm R_i$ and $\bm S_i$. If it is disentangled, its color or language mode's gradients are separately computed from (added or separated) $\frac{\partial \mathcal{L}}{\partial \alpha_i}$. If not, its gradients are added by both color and language modes.

Results shown in Table~\ref{tab:study_disent} validates that disentangling $\bm R$, $\bm S$, and $\alpha$ works the best to preserve the highest image quality. Comparing the fourth and fifth rows, one can also find disentangling $\{\alpha, \bm S\}$ has better effects than $\{\alpha, \bm R\}$. 
This echoes the observation in the main paper Fig.~\ref{fig:RGBL_Gaussain}: language mode prefers larger scales to cover more areas that belong to the same language codes, compared with color rendering that needs smaller Guassians to represent finer textures. Thus, disentangling $\bm S$ shows better performances. 
As a further discussion, disentangling the world-coordinate 3D mean $\bm \mu$ will produce much more Gaussians that attempt to fit in color and language views separately. The setting consumes ~68\% more memory than disentangling $\bm R$, $\bm S$, and $\alpha$ and cannot finish training on a Replica sequence on a RTX-3090 GPU.

\input{sec/4_distangle_room0}

\section{More 3D Localization Evaluation}
\label{supp:3d:loc}

\begin{figure*}[t]
    \centering
    \includegraphics[width=1.0\linewidth]{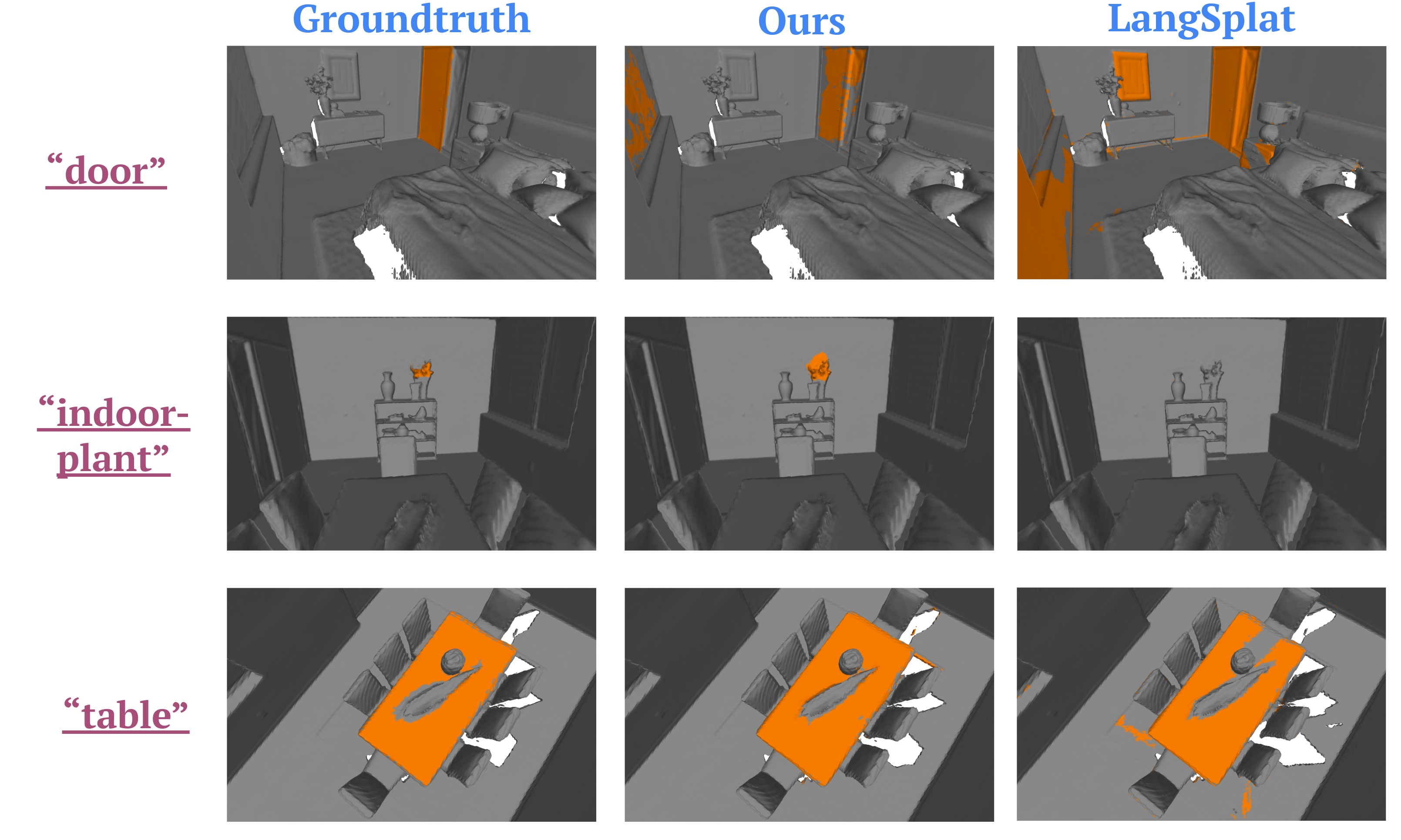}
    \includegraphics[width=1.0\linewidth]{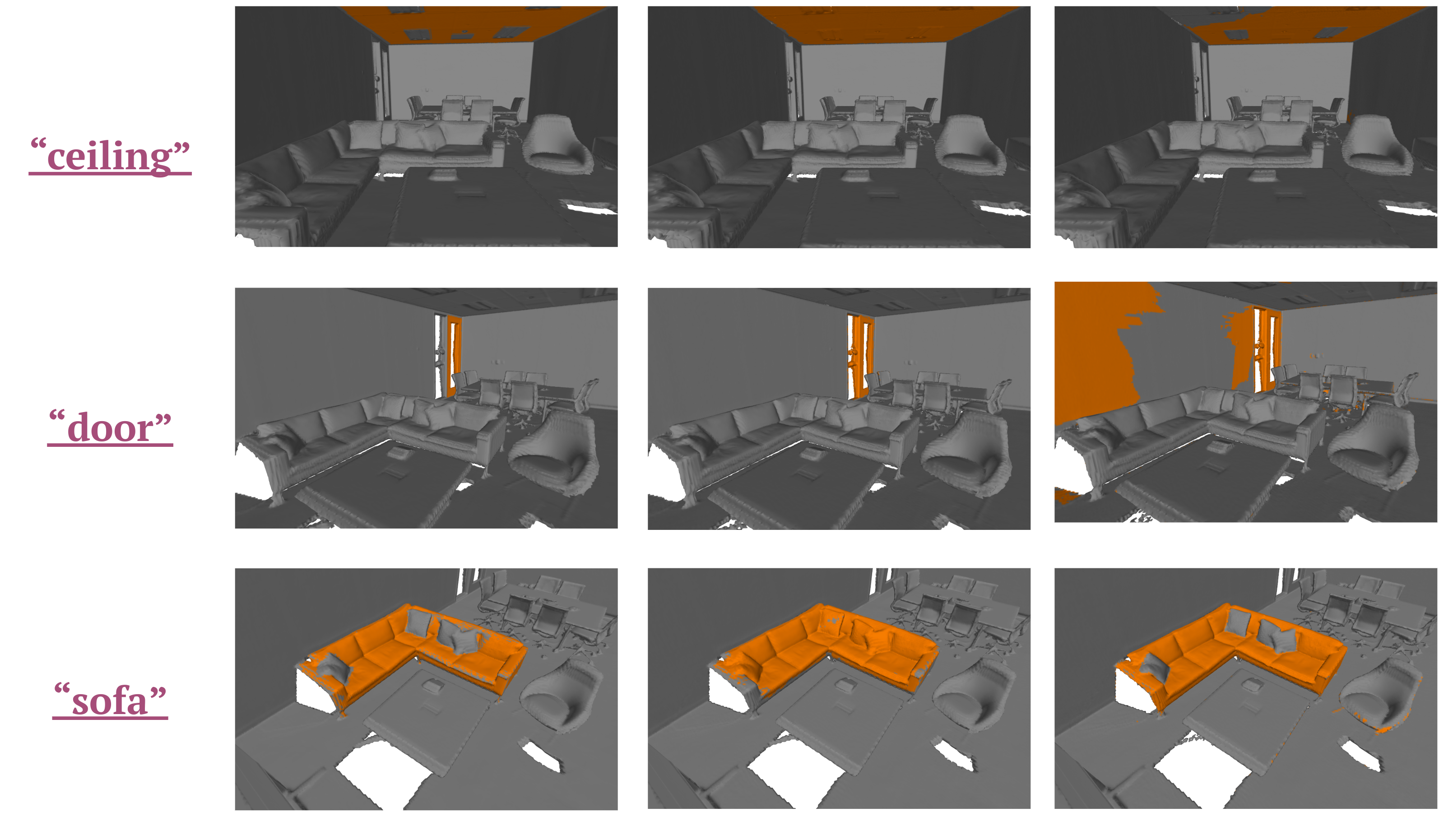}
    \caption{Comparison of 3D localization by queries on Replica sequences.}
    \label{fig:supp_3d_eval_t1}
\end{figure*}

\begin{figure*}[t]
    \centering
    \includegraphics[width=1.0\linewidth]{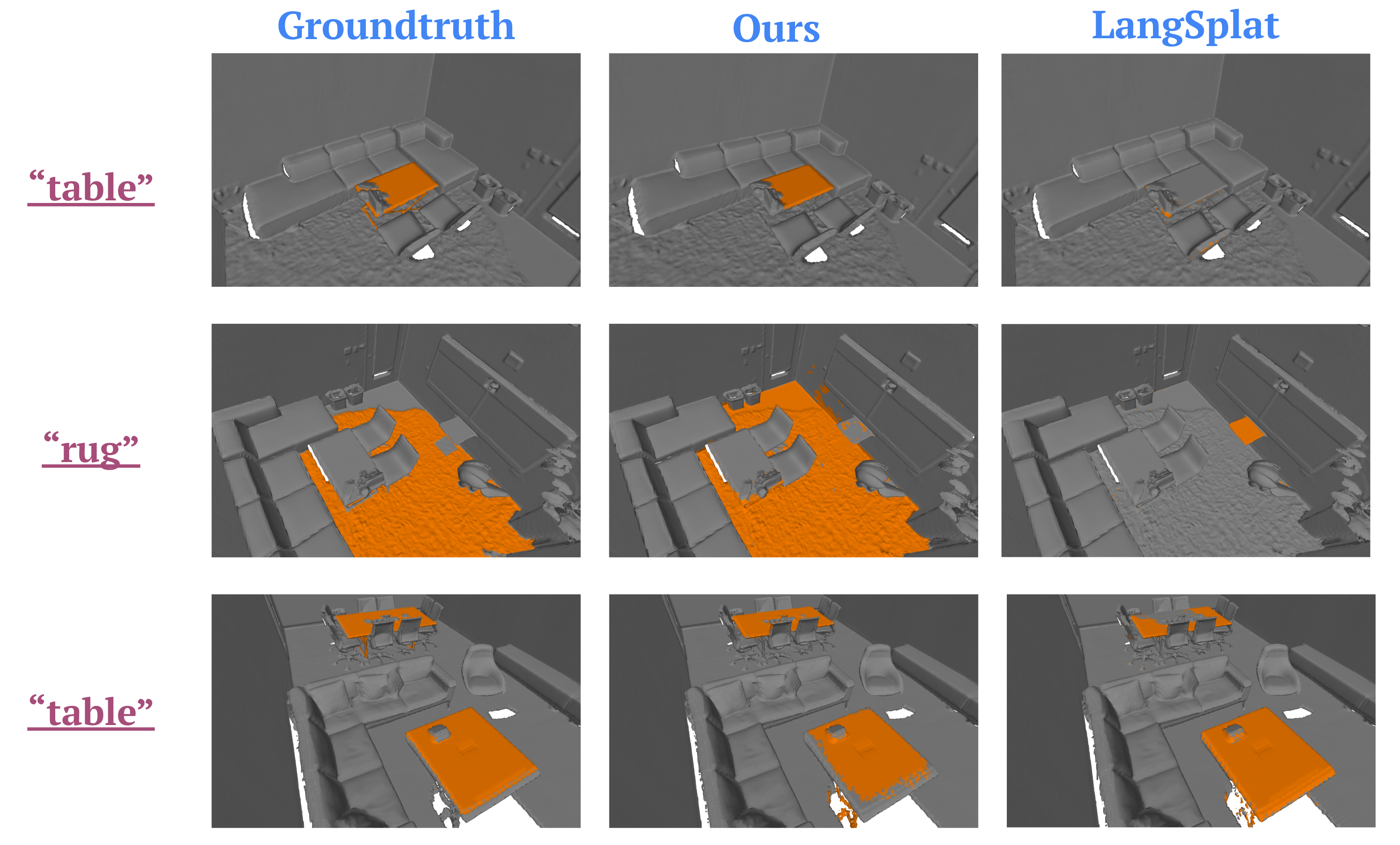}
    \hspace*{-0.5cm}\includegraphics[trim=10mm 0mm 0mm 0mm, clip, width=1.0\linewidth]{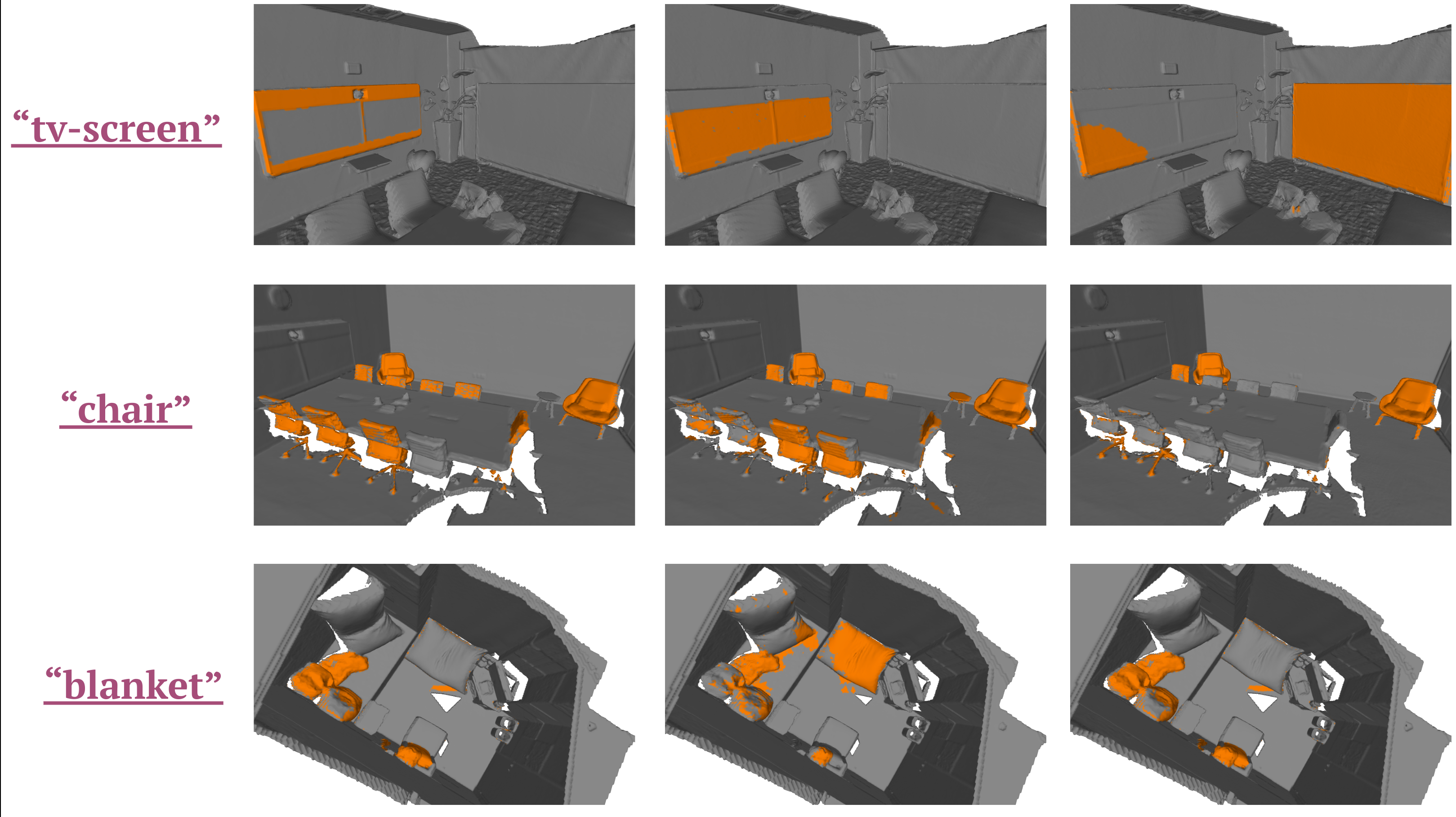}
    \caption{(Continue) Comparison of 3D localization by queries Replica sequences.}
    \label{fig:supp_3d_eval_t2}
\end{figure*}

Following the 3D localization experiment in the main paper, we show more 3D localization evaluation on the 8 sequences of the Replica Dataset. We adopt top-10 frequent categories in each sequence including objects and area, counted by visible pixels. Quantitative results are provided in Table~\ref{tab:3d_eval_all}. 
Note that language label ambiguity exists in Replica's annotations. 
For instance, in the "tv-screen" example in Fig.~\ref{fig:supp_3d_eval_t2}, the groundtruth only counts in the border areas and leaves out the center display areas as "undefined". 
Another example is in Fig.~\ref{fig:supp_3d_eval_t1} the ceiling areas exclude the lights, which is also ambiguous in defining the ceiling regions.
This can result in significantly larger point cloud distances, as measured by the CD and EMD metrics, when queries return objects that do not align with the definitions of the annotation system.
To alleviate this, for CD and EMD, we set a threshold that directly counts a query as failure, when a distance exceeds the metric's population mean plus 2$\times$ population standard deviation. The failures are excluded from the average and reported separately as counts of failure. Results are shown in Table~\ref{tab:3d_eval_all}. Our method performs better than LangSplat on overall CD/ EMD with equal or smaller failure counts.

In Fig.~\ref{fig:supp_3d_eval_t1} and ~\ref{fig:supp_3d_eval_t2}, we visualize more 3D localization results by language queries, which extend Fig.~\ref{fig:3d_localization} in the main paper with the same TSDF procedure to reconstruct meshes.

\section{Limitations and Future Work}
\label{sec:limitation}
This work focuses on static scenes, which may limit its applicability to dynamic environments where objects or spatial configurations change over time. Additionally, both our method and LangSplat are susceptible to false positives for objects that are visually or semantically similar. We notice, for smaller objects SAM generated masks tends to produce more crisp results, whereas for room-sized objects our method shows better localization accuracy due to globally trained pixel-wise CLIP embedding. 

In future work, we aim to extend our approach to dynamic scenes by incorporating mechanisms to handle temporal changes and object motion. Additionally, we plan to explore uncertainty quantification for language features to better evaluate and communicate the reliability of predictions. This improvement would enhance practical use cases, such as robotic navigation and interaction, where confidence in localization is critical.

\end{document}

%% file: sec/4_2_tumrgb_table.tex


\begin{table}[]
\centering
\caption{\textbf{Comparison to Lang-GS SoTA on TUM RGB-D.} Our method is compared to the Lang-GS SoTA method LangSplat on image-based localization accuracy and running time. 
}
\vspace{-5pt}
\begin{footnotesize}
\label{tab:tum_rgbd_comparison}
\setlength{\tabcolsep}{4pt}  
\renewcommand{\arraystretch}{1.0}  
\begin{tabular}{l|cc|cc|c}
\hline
\rowcolor{gray!10}
\textbf{TUM RGB-D} & \multicolumn{2}{c|}{\textbf{Scene1}} & \multicolumn{2}{c|}{\textbf{Scene2}} & \textbf{Time}\\
\cline{2-5}
\rowcolor{gray!10}
 & \textbf{mIOU} & \textbf{Loc} & \textbf{mIOU} & \textbf{Loc} & \\
\hline
LangSplat~\cite{qin2024langsplat}     & \textbf{0.646} & 0.850 & \textbf{0.538} & 0.7825 & 2.1 min/fr \\
Ours & 0.599 & \textbf{0.917} & 0.535 & \textbf{0.7905} & \textbf{0.6 s/fr} \\
\hline
\end{tabular}
\vspace{-8pt}
\end{footnotesize}
\end{table}

%% file: sec/4_langsplatcompare_table.tex





\begin{table}[ht]
\centering
\begin{footnotesize}
\caption{\textbf{3D Localization Evaluation.} Our online method is compared to LangSplat in 3D localization from language query. }
\vspace{-8pt}
\label{tab:langsplat_comp}
\setlength{\tabcolsep}{1pt}  
\begin{tabular}{c|c|llllllll}
\hline
\rowcolor{gray!10}
\multirow{1}{*}{\textbf{Method}} & \multirow{1}{*}{\textbf{Online}}  & \multicolumn{8}{c}{\textbf{3D Language Localization}}  \\
\cline{3-10}
\rowcolor{gray!10}
& & \multicolumn{2}{c|}{cabinet} & \multicolumn{2}{c|}{cushion} & \multicolumn{2}{c|}{stool} & \multicolumn{2}{c}{rug} \\
\rowcolor{gray!10}
& & CD$\downarrow$ & \multicolumn{1}{c|}{EMD$\downarrow$} & CD$\downarrow$ & \multicolumn{1}{c|}{EMD$\downarrow$} & CD$\downarrow$ & \multicolumn{1}{c|}{EMD$\downarrow$} & CD$\downarrow$ & EMD$\downarrow$ \\
\hline
LangSplat & \xmark & \textbf{0.03} & \textbf{0.02} & \textbf{0.23} & 0.10 & \textbf{0.16} & \textbf{0.13} & 0.26 & \textbf{0.38} \\
Ours & \cmark & 0.04 & 0.05 & 0.55 & \textbf{0.03} & 0.20 & 0.40 & \textbf{0.26} & 1.49 \\
\hline
\end{tabular}

\vspace{2mm} 

\begin{tabular}{c|c|llllllll}
\hline
\rowcolor{gray!10}
\multirow{1}{*}{\textbf{Method}} & \multirow{1}{*}{\textbf{Online}}  & \multicolumn{8}{c}{\textbf{3D Language Localization}}  \\
\cline{3-10}
\rowcolor{gray!10}
& & \multicolumn{2}{c|}{lamp} & \multicolumn{2}{c|}{wall} & \multicolumn{2}{c|}{ceiling} & \multicolumn{2}{c}{\textbf{Average}}  \\
\rowcolor{gray!10}
& & CD$\downarrow$ & \multicolumn{1}{c|}{EMD$\downarrow$} & CD$\downarrow$ & \multicolumn{1}{c|}{EMD$\downarrow$} & CD$\downarrow$ & \multicolumn{1}{c|}{EMD$\downarrow$} & CD$\downarrow$ & EMD$\downarrow$   \\
\hline
LangSplat & \xmark & 1.64 & 38.3 & 0.14 & 0.18 & 0.55 & \textbf{0.30} & \cellcolor{thirdorange}0.43 & \cellcolor{thirdorange}5.63  \\
Ours & \cmark & \textbf{1.22} & \textbf{4.08} & \textbf{0.09} & \textbf{0.05} & \textbf{0.27} & 0.71 & \cellcolor{thirdorange}\textbf{0.38} & \cellcolor{thirdorange}\textbf{0.97} \\
\hline
\end{tabular}

\vspace{-2mm}
\end{footnotesize}
\end{table}

%% file: sec/4_1_replica_table.tex
\begin{table*}[t]
\centering
\footnotesize
\caption{\textbf{Comprehensive evaluation on language mapping quality across Replica scenes}. Our method is evaluated against offline SoTA Lang-GS methods on the Replica dataset. We also analyze the impact of our key modules: Super-Resolution Decoder (SRD) and Online Learned AutoEncoder (OLAE) in CLIP Compression. Specifically, for versions requiring in-domain fine-tuning, two scenes from each column are held as testing scenes, while the remaining scenes are used for training.  \colorbox{brightred}{best}, \colorbox{lightred}{second-best}]}
\label{tab:replica_results}
\setlength{\tabcolsep}{4.0pt}
\begin{tabular}{l|cc|cc|cc|cc|cc|c}
\hline
\rowcolor{gray!10}
\textbf{Method} & \multicolumn{2}{c|}{\textbf{Modules}} & \multicolumn{2}{c|}{\textbf{Room0}} & \multicolumn{2}{c|}{\textbf{Room1}} & \multicolumn{2}{c|}{\textbf{Room2}} & \multicolumn{2}{c|}{\textbf{Office0}} & \textbf{Time} \\
\cline{2-11}
\rowcolor{gray!10}
& SRD & OLAE & mIOU & Loc & mIOU & Loc & mIOU & Loc & mIOU & Loc & \\
\hline
LangSplat~\cite{qin2024langsplat} & $-$ & $-$ & 0.356 & 0.710 & 0.393 & 0.718 & 0.413 & 0.694 & 0.353 & 0.526 & 2.8 m/fr \\
Feature3DGS~\cite{zhou2024feature3dgs} & $-$ & $-$ & \colorbox{lightred}{0.487} & 0.677 & 0.301 & 0.812 & 0.353 & 0.800 & 0.342 & 0.661 & 2.3 m/fr \\
LEGaussian~\cite{shi2024language} & $-$ & $-$ & 0.346 & \colorbox{lightred}{0.801} & 0.259 & 0.544 & 0.270 & 0.662 & 0.082 & 0.651 & \colorbox{lightred}{32.1 s/fr} \\
\midrule
\multirow{5}{*}{Ours} 
& \xmark & \xmark & 0.320 & 0.716 & 0.498 & 0.838 & 0.405 & 0.760 & 0.397 & 0.761 & \multirow{5}{*}{\colorbox{brightred}{0.8 s/fr}} \\
& COCO & \xmark & 0.405 & 0.788 & \colorbox{brightred}{0.554} & 0.850 & 0.497 & 0.832 & \colorbox{brightred}{0.457} & \colorbox{brightred}{0.805} &  \\
& COCO & \cmark & 0.389 & 0.773 & 0.493 & 0.832 & \colorbox{brightred}{0.576} & \colorbox{lightred}{0.833} & \colorbox{lightred}{0.454} & 0.758 &  \\
 & Omni & \xmark & 0.414 & 0.706 & 0.499 & \colorbox{lightred}{0.876} & 0.534 & \colorbox{brightred}{0.860} & 0.405 & 0.737 &  \\
 & Omni & \cmark & \colorbox{brightred}{0.552} & \colorbox{brightred}{0.810} & \colorbox{lightred}{0.505} & \colorbox{brightred}{0.939} & \colorbox{lightred}{0.493} & 0.824 & 0.433 & \colorbox{lightred}{0.774} &  \\
\midrule
\rowcolor{gray!10}
\textbf{Method} & \multicolumn{2}{c|}{\textbf{Modules}} & \multicolumn{2}{c|}{\textbf{Office1}} & \multicolumn{2}{c|}{\textbf{Office2}} & \multicolumn{2}{c|}{\textbf{Office3}} & \multicolumn{2}{c|}{\textbf{Office4}} & \textbf{Time} \\
\cline{2-11}
\rowcolor{gray!10}
& SRD & OLAE & mIOU & Loc & mIOU & Loc & mIOU & Loc & mIOU & Loc & \\
\hline
LangSplat~\cite{qin2024langsplat} & $-$ & $-$ & 0.345 & 0.648 & 0.436 & 0.773 & 0.411 & 0.776 & 0.433 & 0.728 & 2.8 m/fr \\
Feature3DGS~\cite{zhou2024feature3dgs} & $-$ & $-$ & 0.254 & 0.495 & 0.387 & \colorbox{lightred}{0.863} & 0.337 & 0.879 & 0.414 & \colorbox{brightred}{0.854} & 2.3 m/fr \\
LEGaussian~\cite{shi2024language} & $-$ & $-$ & 0.354 & 0.414 & 0.178 & 0.680 & 0.267 & \colorbox{brightred}{0.943} & 0.204 & 0.766 & \colorbox{lightred}{32.1 s/fr} \\
\midrule
\multirow{5}{*}{Ours} 
& \xmark & \xmark & 0.219 & 0.502 & 0.450 & 0.830 & 0.481 & 0.838 & 0.431 & 0.790 & \multirow{5}{*}{\colorbox{brightred}{0.8 s/fr}} \\
& COCO & \xmark & 0.272 & 0.393 & 0.570 & 0.847 & \colorbox{lightred}{0.553} & \colorbox{lightred}{0.919} & \colorbox{lightred}{0.492} & \colorbox{lightred}{0.824} &  \\
& COCO & \cmark & \colorbox{brightred}{0.357} & 0.525 & \colorbox{lightred}{0.574} & 0.820 & 0.495 & 0.766 & \colorbox{brightred}{0.498} & 0.765 &  \\
 & Omni & \xmark & 0.388 & \colorbox{lightred}{0.674} & \colorbox{brightred}{0.610} & \colorbox{brightred}{0.889} & 0.455 & 0.802 & 0.455 & 0.802 &  \\
 & Omni & \cmark & \colorbox{brightred}{0.357} & \colorbox{brightred}{0.734} & 0.522 & 0.826 & \colorbox{brightred}{0.578} & 0.887 & 0.458 & 0.812 &  \\
\hline
\end{tabular}

\end{table*}

%% file: sec/4_slam_sota_table.tex
\begin{table*}[t]
\centering
\begin{footnotesize}
\caption{\textbf{Per Scene Evaluation of SLAM-3DGS on Replica}. Our method is evaluated against other SLAM-3DGS approaches based on novel view rendering quality and camera localization error (ATE in cm). [Key: \colorbox{brightred}{best}, \colorbox{lightred}{second-best}]}
\vspace{-7pt}
\label{tab:SOTA_comparison_slam_gs}
\setlength{\tabcolsep}{1pt}  
\begin{tabular}{l|c|cccc|cccc|cccc|cccc}
\hline
\rowcolor{gray!10}
\multirow{1}{*}{\textbf{Method}} & \multirow{1}{*}{\textbf{w/ Lang.}} & \multicolumn{4}{c|}{\textbf{Room0}} & \multicolumn{4}{c|}{\textbf{Room1}} & \multicolumn{4}{c|}{\textbf{Room2}} & \multicolumn{4}{c}{\textbf{Office0}} \\
\cline{3-18}
\rowcolor{gray!10}
& & PSNR↑ & SSIM↑ & LPIPS↓ & ATE↓ & PSNR↑ & SSIM↑ & LPIPS↓ & ATE↓ & PSNR↑ & SSIM↑ & LPIPS↓ & ATE↓ & PSNR↑ & SSIM↑ & LPIPS↓ & ATE↓ \\
\hline
SplaTAM~\cite{keetha2024splatam} & \xmark & 32.31 & \colorbox{brightred}{0.974} & \colorbox{brightred}{0.072} & 0.47 & 33.36 & \colorbox{lightred}{0.966} & 0.101 & 0.42 & \colorbox{lightred}{34.78} &  \colorbox{brightred}{0.983} & \colorbox{brightred}{0.073} & 0.32 & 38.16 & \colorbox{lightred}{0.982} & 0.084 & 0.46 \\
RTG-SLAM~\cite{peng2024rtg} & \xmark & 31.56 & \colorbox{lightred}{0.967} & 0.131 & 0.20 & \colorbox{brightred}{34.21} & \colorbox{brightred}{0.979} & 0.105 & 0.18 & \colorbox{brightred}{35.57} & \colorbox{lightred}{0.981} & 0.115 & 0.13 & \colorbox{lightred}{39.11} & \colorbox{brightred}{0.990} & 0.068 & 0.22\\
MonoGS~\cite{matsuki2024gaussian} & \xmark & \colorbox{lightred}{33.36} & 0.941 & 0.086 & 0.458 &\colorbox{lightred}{33.58} & 0.942 & \colorbox{lightred}{0.086} & 0.424 & 34.12 & 0.950 & 0.081 & 0.490 &\colorbox{brightred}{40.91} & 0.980 & \colorbox{brightred}{0.045} & 0.615\\
Ours & \cmark  & \colorbox{brightred}{33.38} & 0.940 & \colorbox{lightred}{0.085} & 0.325 & 33.46 & 0.941 & \colorbox{brightred}{0.079} & 0.416 & 34.35 & 0.952 & \colorbox{lightred}{0.075} & 0.483 & \colorbox{brightred}{40.91} & 0.978 & \colorbox{lightred}{0.048} & 0.550\\
\midrule
\rowcolor{gray!10}
\multirow{1}{*}{\textbf{Method}} & \multirow{1}{*}{\textbf{w/ Lang.}} & \multicolumn{4}{c|}{\textbf{Office1}} & \multicolumn{4}{c|}{\textbf{Office2}} & \multicolumn{4}{c|}{\textbf{Office3}} & \multicolumn{4}{c}{\textbf{Office4}} \\
\cline{3-18}
\rowcolor{gray!10}
& & PSNR↑ & SSIM↑ & LPIPS↓ & ATE↓ & PSNR↑ & SSIM↑ & LPIPS↓ & ATE↓ & PSNR↑ & SSIM↑ & LPIPS↓ & ATE↓ & PSNR↑ & SSIM↑ & LPIPS↓ & ATE↓ \\
\hline
SplaTAM~\cite{keetha2024splatam} & \xmark & 38.49 & \colorbox{lightred}{0.980} & 0.095 & 0.24 & 31.66 & \colorbox{lightred}{0.962} & \colorbox{brightred}{0.102} & 0.28 & 29.24 & 0.948 & 0.123 & 0.39 & 31.54 & 0.946 & 0.157 & 0.56 \\
RTG-SLAM~\cite{peng2024rtg} & \xmark & \colorbox{brightred}{40.24} & \colorbox{brightred}{0.992} & 0.075 & 0.12 & \colorbox{lightred}{33.54} & \colorbox{brightred}{0.981} & 0.128 & 0.22 & \colorbox{brightred}{36.48} & \colorbox{brightred}{0.984} & 0.117 & 0.20 & \colorbox{lightred}{35.43} & \colorbox{brightred}{0.982} & 0.109 & 0.19 \\
MonoGS~\cite{matsuki2024gaussian} & \xmark & \colorbox{lightred}{39.77} & 0.976 & \colorbox{lightred}{0.049} & 0.327 & \colorbox{brightred}{33.81} & 0.907 & \colorbox{lightred}{0.114} & 0.341 & \colorbox{lightred}{35.17} & 0.954 & \colorbox{lightred}{0.058} & 0.303 & 35.02 & 0.952 & \colorbox{lightred}{0.082} & 0.405 \\
Ours & \cmark & 39.60 & 0.976 & \colorbox{brightred}{0.044} & 0.382 & 33.05 & 0.901 & 0.125 & 0.396 & 34.98 & \colorbox{lightred}{0.955} & \colorbox{brightred}{0.053} & 0.203 & \colorbox{brightred}{36.75} & \colorbox{lightred}{0.957} & \colorbox{brightred}{0.063} & 0.423 \\
\hline
\end{tabular}

\end{footnotesize}
\end{table*}

%% file: sec/4_distangle_room0.tex
\begin{table*}[ht]
\centering
\begin{footnotesize}
\caption{\textbf{Effects of Disentangled GS Parameters (Category-Wise)}. Here, we show category-wise results for language-queried 3D localization on the Replica Room-0 subset.}
\vspace{-8pt}
\label{tab:disentangledGS}
\setlength\tabcolsep{2pt}
\begin{tabular}
{l|l l l l l l l l l l l l l l l l}
\hline
\rowcolor{gray!10}
\multirow{1}{*}{\textbf{Method}} & \multicolumn{16}{c}{\textbf{3D Language Query}}  \\
\cline{2-17}
\rowcolor{gray!10}
& \multicolumn{2}{c|}{cabinet} & \multicolumn{2}{c|}{cushion} & \multicolumn{2}{c|}{stool} & \multicolumn{2}{c|}{rug} & \multicolumn{2}{c|}{lamp} & \multicolumn{2}{c|}{wall} & \multicolumn{2}{c|}{ceiling} & \multicolumn{2}{c}{\textbf{Average}} \\
\rowcolor{gray!10}
& CD $\downarrow$ & \multicolumn{1}{c|}{EMD$\downarrow$} & CD $\downarrow$ & \multicolumn{1}{c|}{EMD$\downarrow$} & CD$\downarrow$ & \multicolumn{1}{c|}{EMD$\downarrow$} & CD$\downarrow$  & \multicolumn{1}{c|}{EMD$\downarrow$} & CD$\downarrow$ & \multicolumn{1}{c|}{EMD$\downarrow$} & CD$\downarrow$ & \multicolumn{1}{c|}{EMD$\downarrow$} & CD$\downarrow$ & \multicolumn{1}{c|}{EMD$\downarrow$} &CD$\downarrow$ & \multicolumn{1}{c}{EMD$\downarrow$} \\
\hline
Joint RGB-L& 0.042 & \textbf{0.019} & 1.252 & 3.022 & \textbf{0.128} & \textbf{0.007} & 0.283 & \textbf{0.866} & \textbf{0.579} & 9.517 & 0.110 & 0.118 & 0.295 & \textbf{0.033} & \cellcolor{thirdorange}0.384 & \cellcolor{thirdorange}1.940 \\
Disentangled & \textbf{0.040} & 0.053 & \textbf{0.554} & \textbf{0.032} & 0.196 & 0.399 & \textbf{0.256} & 1.494 & 1.222 & \textbf{4.083} & \textbf{0.087} & \textbf{0.050} & \textbf{0.269} & 0.707 & \cellcolor{thirdorange}\textbf{0.375} & \cellcolor{thirdorange}\textbf{0.974}  \\
\hline
\end{tabular}
\vspace{-2mm}
\end{footnotesize}
\end{table*}